\pdfoutput=1
\PassOptionsToPackage{table}{xcolor}
\documentclass[11pt]{article}

\usepackage[]{acl}

\usepackage{times}
\usepackage{latexsym}

\usepackage[T1]{fontenc}

\usepackage[utf8]{inputenc}

\usepackage{microtype}

\usepackage{amsmath}
\usepackage{amssymb}
\usepackage{booktabs} 
\usepackage{mathtools}
\usepackage{amsthm}
\usepackage{threeparttable}
\usepackage[capitalize,noabbrev]{cleveref}
\usepackage{subcaption}
\usepackage{graphicx}
\usepackage{float}
\usepackage{xcolor}
\usepackage{fontawesome}
\definecolor{brandeisblue}{rgb}{0.0, 0.44, 1.0}
\definecolor{mygray}{RGB}{220,220,220}

\newcommand{\ignore}[1]{}

\usepackage{snaptodo}
\snaptodoset{block rise=2em}
\snaptodoset{margin block/.style={font=\tiny}}

\title{Reconfidencing LLM Uncertainty from the Grouping Loss Perspective}

\author{
Lihu Chen\textsuperscript{\rm 1,3},
Alexandre Perez-Lebel\textsuperscript{\rm 1}, 
Fabian M. Suchanek\textsuperscript{\rm 2},
Gaël Varoquaux\textsuperscript{\rm 1}\\
\textsuperscript{\rm 1} Soda, Inria Saclay, France \\
\textsuperscript{\rm 2} LTCI, Télécom Paris, Institut Polytechnique de Paris, France \\
\textsuperscript{\rm 3} Imperial College London, UK \\
\texttt{\{lihu.chen\}@imperial.ac.uk}\\
\texttt{\{alexandre.perez, gael.varoquaux\}@inria.fr}\\ 
\texttt{\{fabian.suchanek\}@telecom-paris.fr}
}

\begin{document}
\maketitle
\begin{abstract}
Large Language Models (LLMs), such as GPT and LLaMA, are susceptible to generating hallucinated answers in a confident tone. 
While previous efforts to elicit and calibrate uncertainty have shown some success, they often overlook biases towards certain groups, such as specific nationalities. Existing calibration methods typically focus on average performance, failing to address this disparity.
In our study, we demonstrate that the concept of grouping loss is an effective metric for understanding and correcting the heterogeneity in confidence levels. 
We introduce a novel evaluation dataset, derived from a knowledge base, specifically designed to assess the confidence scores of LLM responses across different groups.
Our experimental results highlight significant variations in confidence, which are accurately captured by grouping loss. To tackle this issue, we propose a new method to calibrate the confidence scores of LLMs by considering different groups, a process we term \emph{reconfidencing}. Our findings indicate that this approach effectively mitigates biases against minority groups, contributing to the development of fairer LLMs.
The code is available at \faGithub~ \url{https://github.com/tigerchen52/reconfidencing_llms}
\end{abstract}

\section{Introduction}
\label{submission}

While Large Language Models (LLMs) such as ChatGPT~\cite{chatgpt} and LLaMA~\cite{touvron2023llama} 
can 
generate responses that are fluent and plausible, 
they can also provide incorrect and untruthful information in a confident and compelling tone. 
This phenomenon, often called hallucination, poses a notable challenge to their use
~\cite{ji2023survey,baan2023uncertainty}. 

\begin{figure}
    \centering
    \includegraphics[width=0.5\textwidth]{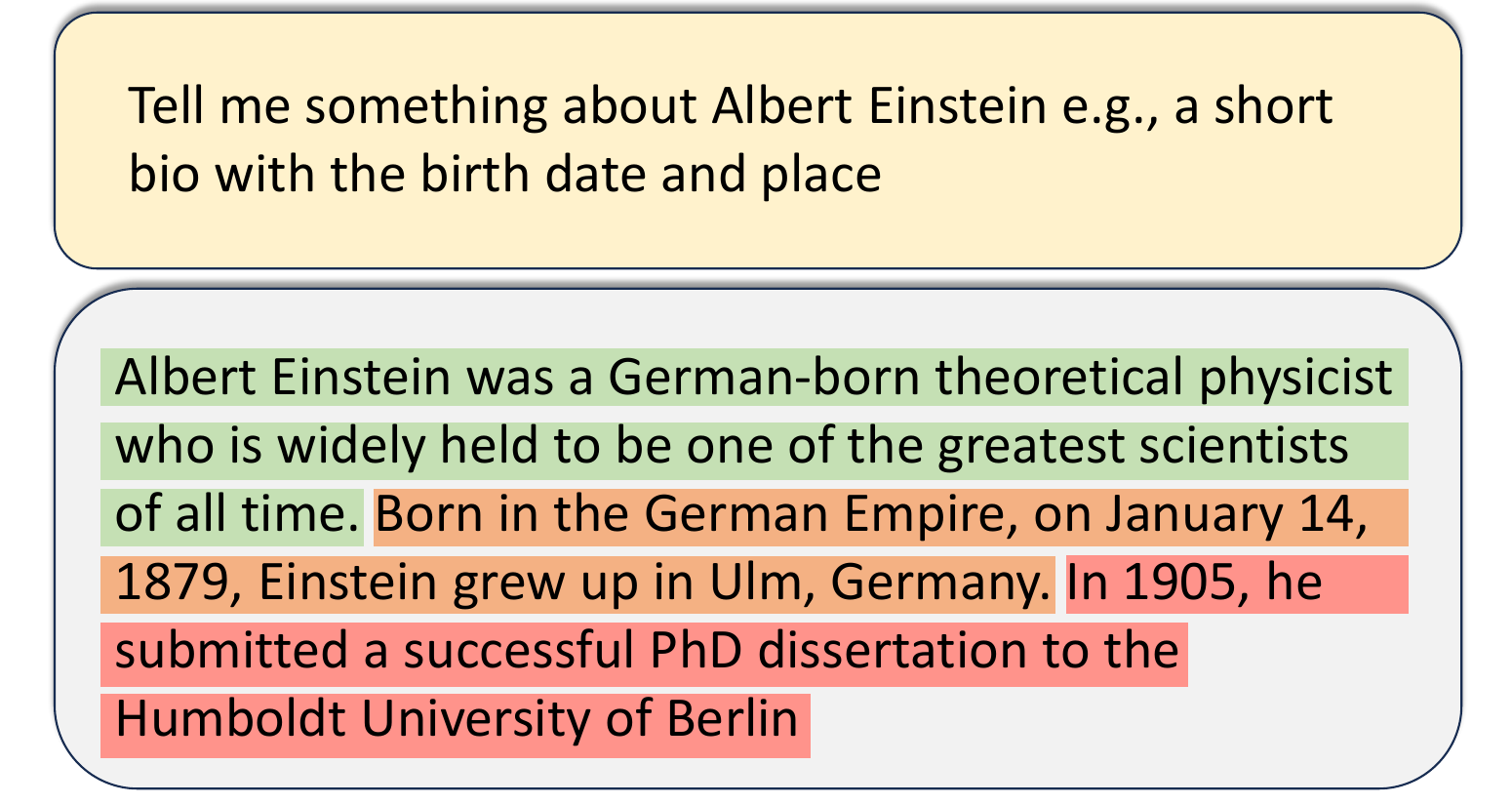}
    \caption{\textbf{Desired user experience} -- An illustration of our goals of eliciting confidence levels in LLMs. High confidence scores are represented in green, while red indicates a higher likelihood of encountering hallucinated sentences.}
    
    \label{fig:confidence_example}
\end{figure}

In response, extensive research has focused on estimating the confidence (or uncertainty) of LLM answers
~\cite{huang2023survey, zhang2023siren}. 
Through expressions of confidence levels,  we know to what degree to trust a statement rather than blindly believing. Figure~\ref{fig:confidence_example} illustrates an ideal user experience, where LLMs document sentence-level confidence in their answers. 
Methods of estimating confidence can be categorized into two groups: \emph{White-box} and \emph{Black-box} methods.  
White-box methods require access to internal states~\cite{azaria2023internal} or model logits~\cite{lin2022teaching} while Black-box methods rely solely on text responses to obtain confidence scores. In cases where the LLM allows only restricted access to internal states (e.g., ChatGPT), black-box methods are more suitable. These methods establish confidence scores 
by analyzing the consistency of multiple answers to a single query~\cite{kuhn2022semantic,manakul2023selfcheckgpt} or by creating specific prompts to capture expressed confidence scores~\cite{zhou2023navigating,xiong2023can,tian2023just}.

\begin{figure*}[t]%
	\centering
	\subfloat[\centering Overall]{{\includegraphics[width=0.32\textwidth]{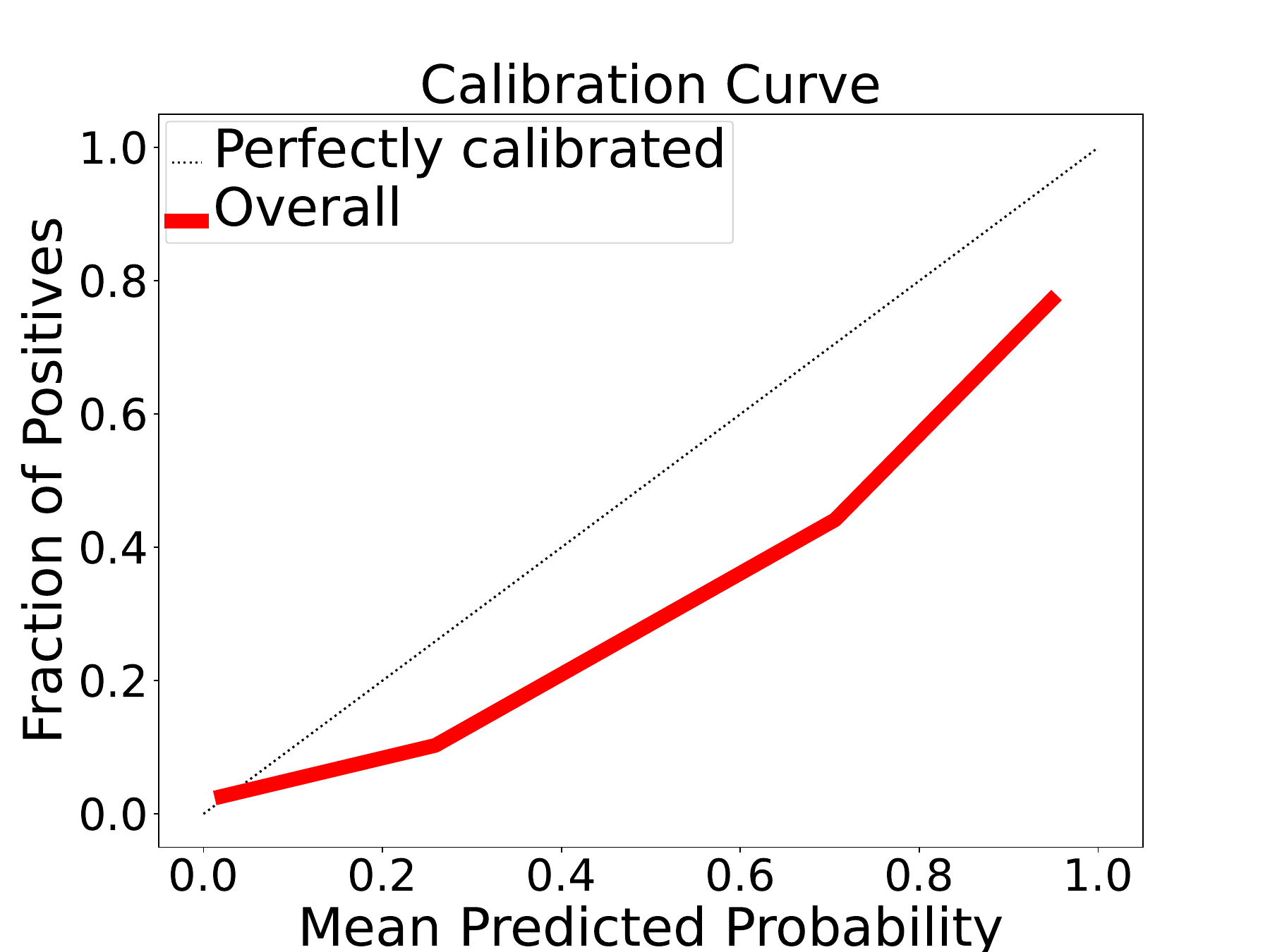} }\label{fig:birth_date_a}}
	\subfloat[\centering Popularity]{{\includegraphics[width=0.32\textwidth]{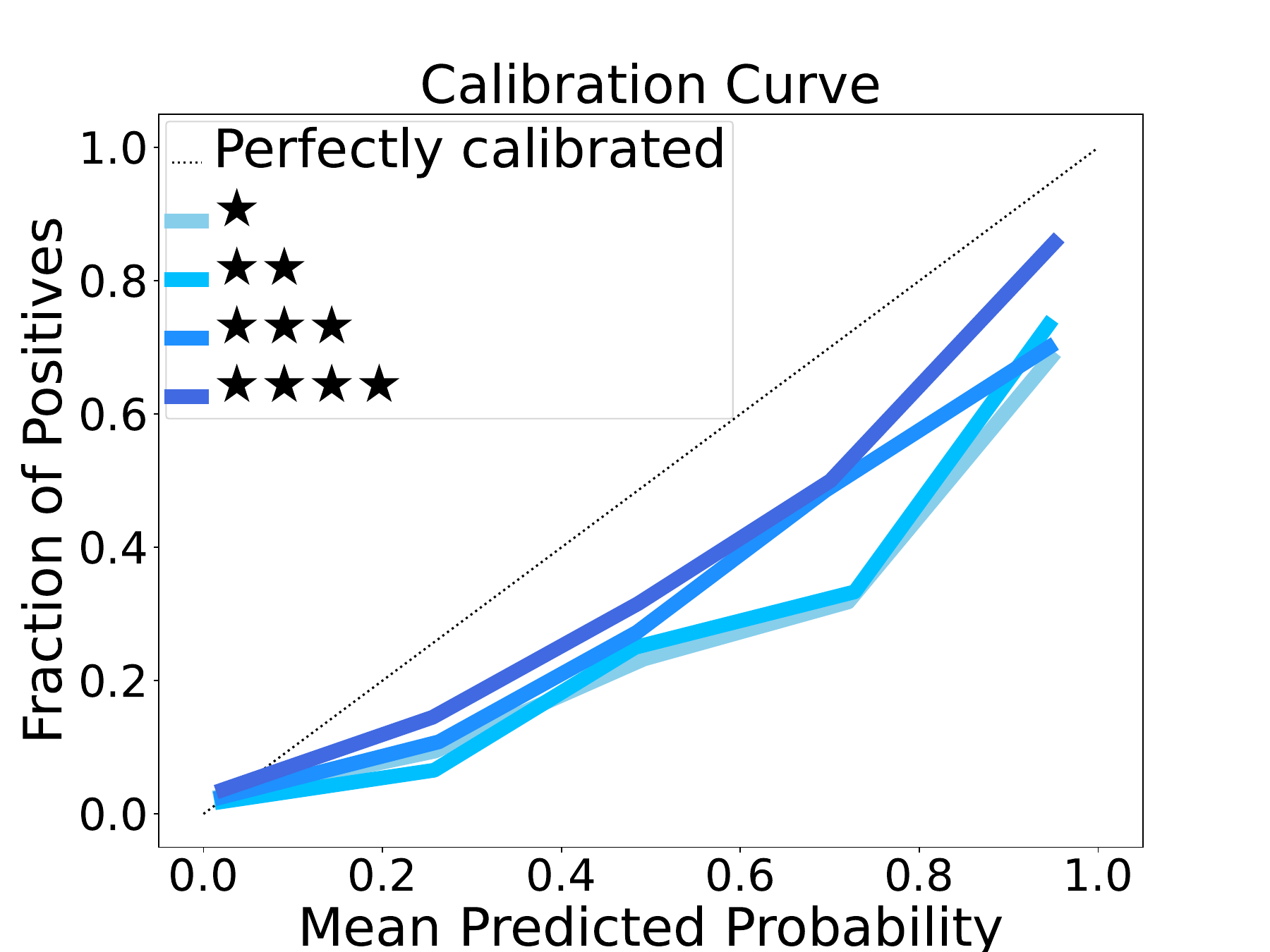} }\label{fig:birth_date_b}}%
 \subfloat[\centering Nationality]{{\includegraphics[width=0.32\textwidth]{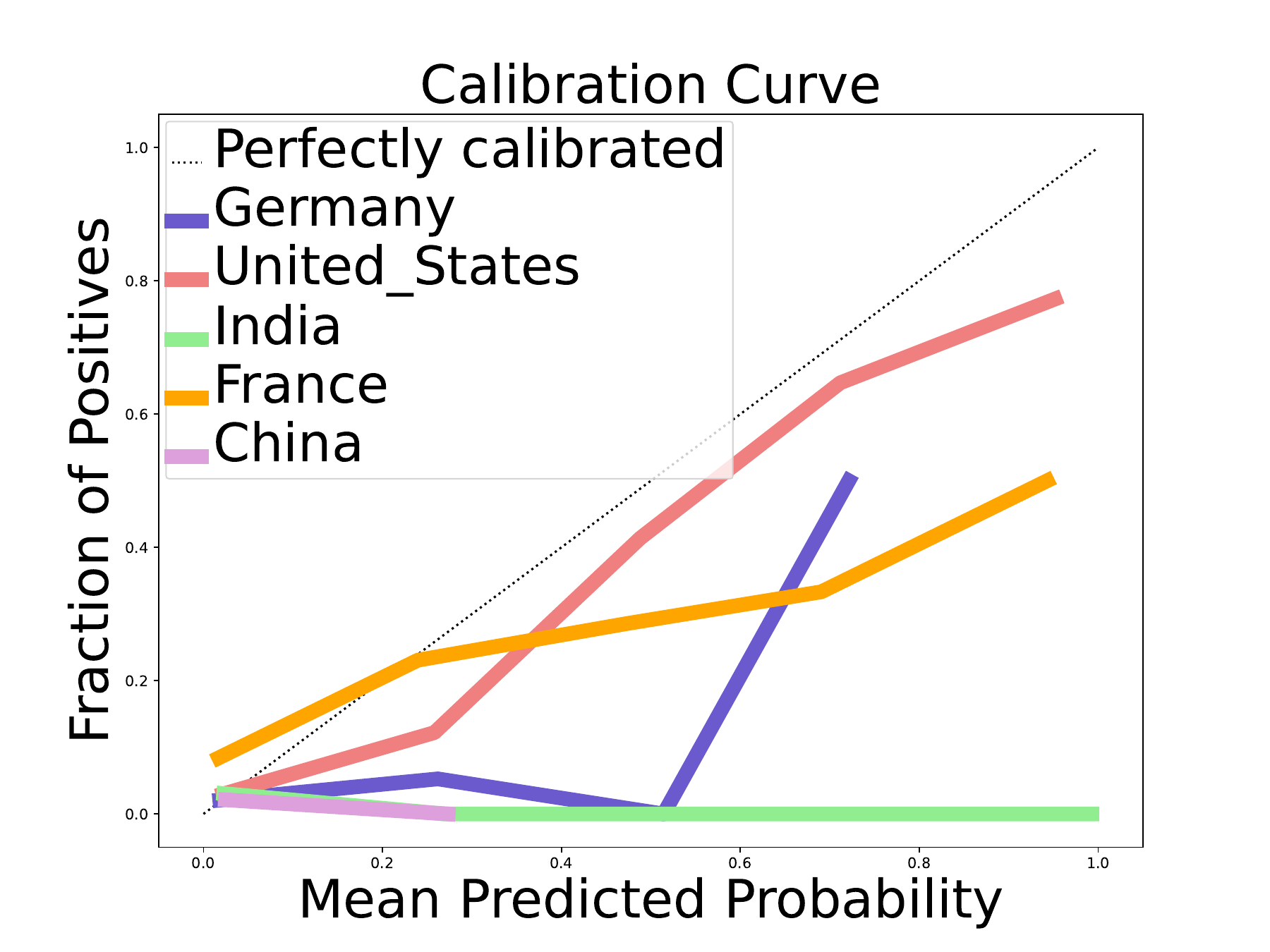} }\label{fig:birth_date_c}}%
	\caption{Calibration curves of the \texttt{Birth\_Date} relation. The LLM here is Mistral-7B~\cite{jiang2023mistral}, and we use SelfCheckGPT~\cite{manakul2023selfcheckgpt} to compute confidence scores.  An increased number of $\star$ symbols signifies a sub-group containing more popular samples.}
	\label{fig:birth_date_example}%
 \vspace{-10pt}
\end{figure*}

Although some methods use calibration to adjust the predictions of a model to better match the true probabilities ~\cite{hendrycks2020measuring, gawlikowski2021survey, mielke2022reducing, tian2023just}, these approaches predominantly concentrate on average performance metrics, often neglecting the heterogeneity among different groups.
Consequently, calibration alone proves inadequate. Even when a calibration technique attains optimal average accuracy, the calibrated scores can still markedly deviate from the true posterior probabilities for specific groups of queries -- a phenomenon known as the grouping loss~\cite{kull2015novel,perez2022beyond}.
As an example, let us consider a query that asks for the birth dates of people, as in
\textit{``What is the birth date of Albert Einstein?''}. We submitted this query for 5K people
to an LLM \citep[Mistral-7B,][]{jiang2023mistral}, and generated a confidence score for each answer with a consistency-based method \citep[SelfCheckGPT,][]{manakul2023selfcheckgpt}.  In a classic calibration analysis, we grouped the answers into buckets by their confidence score, and computed the observed ratio of correct answers in each bucket. 
Figure~\ref{fig:birth_date_a} shows the corresponding calibration curve for all test samples. The curve is close to the diagonal, which means that the confidence score is close to the true ratio of correct answers in each bucket.
This picture changes a bit when we split our data into popular and less popular persons based on the backlink numbers. As shown in Figure~\ref{fig:birth_date_b}, answers on more popular entities tend to be better calibrated than answers on long-tail entities.
The picture is even more dramatic when we split the people by nationality (Figure~\ref{fig:birth_date_c}):
While the calibration is satisfactory for American and French individuals, it performs dismally for almost all Indian and Chinese people. This illustrates grouping loss: a model's calibration error may be small overall, but can be catastrophically large for certain sub-groups.
A well-calibrated LLM might be biased, generating with high confidence untruthful information about a particular race, gender, etc.

In this paper, we conduct a systematic study to measure the error of the confidence estimations.
We create a new dataset that enables evaluating the quality of confidence scores for different types of groups. Our dataset consists of questions about entities (people, locations, etc.) and the ground truth from the YAGO knowledge base~\cite{suchanek2023integrating}. 
In addition, our dataset contains features of the entities, such as popularity and nationality, which allows us to study sub-groups of entities.
We evaluate two recently proposed methods for deriving confidence levels: \textit{SelfCheckGPT}~\cite{manakul2023selfcheckgpt} and \textit{Just Ask for Calibration}~\cite{tian2023just}.
To identify grouping loss, we use both user-defined 
and latent groups. User-defined groups rely on features (which may be hand-crafted) such as popularity and nationality, while latent groups are automatically identified by decision trees~\cite{perez2022beyond}. 
Experiments reveal that models like Mistral and LLaMA tend to be overly confident across all questions. In addition, they are more confident on some queries than others: they display grouping loss.
To improve confidence scores, we propose an approach to adjust LLMs, tackling both calibration and grouping loss. The core idea is to calibrate the confidence score for each sub-group separately, a method we term \emph{reconfidencing}.
Experimental results show that our refined solution has a better performance in terms of Brier score and grouping loss.

In summary, our contributions are threefold:
\begin{itemize}
    \item We introduce a new framework and dataset to analyze the capability of LLMs to elicit confidence scores for different groups
    \item We prove the existence of the grouping loss in LLMs and compare the heterogeneity of confidence errors on both user-defined groups and implicit groups
    \item We propose a refined way to reconfidence LLMs from a group-level perspective, which can reduce discrimination of minority groups and lead to fairer LLMs. 
\end{itemize}

\section{Related Work}

\subsection{Confidence Elicitation in LLMs}
To alleviate the hallucination phenomenon, some methods attempt to elicit confidence (or uncertainty) scores for the generated answers of LLMs~\cite{ji2023survey, zhang2023siren, huang2023survey}. 
These efforts can be roughly categorized into two groups: \emph{White-box} and \emph{Black-box} methods.
White-box methods need access to internal states or token logits while Black-box methods use only textual responses to compute confidence scores. 

There are three primary white-box ways to encourage LLMs to express uncertainty in a human-like manner: Verbalized Probability, Internal State, and Token Logit.
The goal of \textit{verbalized probability} is to teach models to convey its degree of certainty, as in \textit{I'm 90\% sure that it is...}. The models are fine-tuned on particular tasks~\cite{lin2022teaching}  to elicit probabilistic responses.
The \textit{internal state method} builds a classifier to detect the truthfulness of a statement, which receives as input the activation values of the hidden layers of an LLM~\cite{azaria2023internal}.
The \textit{token logit method} evaluates the probability distribution of the words in the answer. At each step, LLMs produce a probability distribution across the entire vocabulary. Analyzing the distribution allows us to compute corresponding entropy values, which serve as indicators of confidence~\cite{fu2023gptscore, manakul2023selfcheckgpt}.
Generally, factual statements tend to feature tokens with higher
likelihood and lower entropy, while hallucinated texts
are likely to come from positions with flat probability distributions with high uncertainty. 

White-box methods need access to internal states or token logits which are unavailable for some LLMs such as ChatGPT. 
In such cases, one can use black-box methods, which rely solely on the textual answers of LLM.
There are three main black box methods. 
The first relies on asking the same question to an LLM multiple times and assessing the coherence of its responses~\cite{kuhn2022semantic, manakul2023selfcheckgpt,lin2023generating,xiong2023can}. If the answers contradict each other, one assumes a lack of confidence in the statement.
The second method uses external resources and tools to verify the answers.
For example, symbolic knowledge bases and search engines can be leveraged to fact-check LLM outputs~\cite{gou2023critic,agrawal2023language}. 
Finally, a third branch of approaches resorts to in-context learning prompts for obtaining confidence scores~\cite{zhou2023navigating, xiong2023can,tian2023just}.  

\subsection{Confidence Calibration and Grouping Loss}
Ideally, a model's confidence score should equal the actual probability of the answer being correct. 
Recent studies have shown that current powerful models are poorly calibrated: they are over-confident or (more seldom) under-confident. This holds both for modern neural networks~\cite{guo2017calibration} and LLMs like GPT~\cite{hendrycks2020measuring}. Dedicated approaches have been proposed to calibrate these models~\cite{gawlikowski2021survey,jiang2021can, park2022calibration, kadavath2022language, xiao2022uncertainty, mielke2022reducing}. 
Yet calibration
is not enough: even a perfectly calibrated classifier
can have confidence scores that are far from the true posterior probabilities for certain types of questions -- 
a phenomenon known as the grouping loss~\cite{kull2015novel}. \citet{perez2022beyond} recently contributed a measure for the grouping loss, which captures heterogeneity in the confidence score. They revealed grouping loss on pre-trained vision and text classifiers, but did not study generative models.
In this work, we are the first to study the grouping loss of generative models.
We are also the first to propose a method to reconfidence LLMs from the grouping loss perspective.

\section{Analyzing the Grouping Loss in LLMs}
In this section, we aim to measure the calibration of existing confidence methods and identify the grouping loss in LLMs.


\subsection{Dataset Construction}~\label{sec: dataset_construction}

To study the grouping loss in LLM confidence scores, we need control over the entities that appear in the questions, to vary their properties and examine calibration errors.

For this purpose, we construct a new evaluation dataset derived from the YAGO knowledge base~\cite{suchanek2023integrating}.
YAGO contains triples of a subject, a relation, and an object, as in $\langle$\textit{Albert Einstein, Birth Date, 1879-03-14}$\rangle$. We select three relations: \texttt{Birth Date}, \texttt{Founder}, and \texttt{Composer}. This choice is driven by the desire to cover different top-level classes (people, organizations, and creative works). Furthermore, these relations have few objects per subject, which makes it very likely that the KB contains the complete list of objects for a given subject~\cite{amieplus}. Finally, the relations cover both functional relations (with one object per subject) and non-functional ones (with potentially several objects per subject). 
We collect around 10 thousand triples for each relation. Each triple comes with a  natural language question that we generate with a template, as in \textit{``What is the birth date of the person Albert Einstein?''}. 

In addition, our dataset contains some hand-picked facts about the subject of each triple such as nationality and gender.
We also store the popularity of an entity, which we obtained by 
 the Backlinks API\footnote{\href{https://www.mediawiki.org/wiki/API:Backlinks}{www.mediawiki.org/wiki/API:Backlinks}. The backlink number shows an entity appears how many times in other Wikipedia pages} and YAGO, respectively. Table~\ref{tab:dataset_stats} shows the statistics of our dataset. 

Since we need to learn decision tree classifiers 
and calibrators in the subsequent experiments, the dataset is split into training, validation, and test sets according to the ratio of 0.25:0.25:0.50. All the following reported scores are based on the test set.

\subsection{Experimental Settings}~\label{sec:experimental_settings}

\paragraph{LLMs.} In this experiment, we focus on instruction-aligned LLMs~\cite{ouyang2022training}, which are widely used in various applications. 
Also, we study open-source models since it is necessary for our method to access internal input representations when reconfidencing LLMs, which we will talk about later.
We consider three open-source LLMs with different sizes: LLaMA~\cite{touvron2023llama}, Mistral~\cite{jiang2023mistral}, and Mixtral~\cite{jiang2024mixtral}, all downloaded from HuggingFace.
Note that our method is model-agnostic and can be applied to other LLMs as well.

\begin{table*}[!t]  
	\centering  
	\scriptsize
        \setlength{\tabcolsep}{3.5mm}{
	\begin{threeparttable} 
		\begin{tabular}{cccccc}  
			\toprule  
			Relation&Size&Head&Tail&Query Example&Answer Example\cr
			\midrule
			\texttt{Birth\_Date}&10,000&Person&Date&\textit{What is the birth year of the person Albert Einstein?} &\textit{1879}\cr 
			\texttt{Founder}&10,000&Business&Person&\textit{Who founded  the business Microsoft?} &\textit{Bill Gates}\cr
                \texttt{Composer}&9,419&Music&Person&\textit{Who composed the song Rolling in the Deep?} &\textit{Adele}\cr
			\bottomrule  
		\end{tabular}
		\caption{Description of our evaluation dataset. Note that there might be multiple answers for the founder and composer relations and we predict only the birth year for the \texttt{Birth\_Date} relation.}\label{tab:dataset_stats}
	\end{threeparttable}  
 }
\end{table*}

\begin{table*}[bt] 
	\centering
        \scriptsize
	\setlength{\tabcolsep}{3.5mm}{
		\begin{threeparttable} 
			\begin{tabular}{c|ccc|ccc|ccc}  
				\toprule
				\textbf{Method}
				&\multicolumn{3}{c|}{\textbf{\underline{\texttt{Birth\_Date}}}}&\multicolumn{3}{c|}{\textbf{\underline{\texttt{Founder}}}}&\multicolumn{3}{c}{\textbf{\underline{\texttt{Composer}}}}\cr
			&Brier~\textcolor{brandeisblue}{$\downarrow$}&CL~\textcolor{brandeisblue}{$\downarrow$}&GL~\textcolor{brandeisblue}{$\downarrow$}&Brier~\textcolor{brandeisblue}{$\downarrow$}&CL~\textcolor{brandeisblue}{$\downarrow$}&GL~\textcolor{brandeisblue}{$\downarrow$}&Brier~\textcolor{brandeisblue}{$\downarrow$}&CL~\textcolor{brandeisblue}{$\downarrow$}&GL~\textcolor{brandeisblue}{$\downarrow$}\cr
			 \midrule
                    \textit{LLaMA-7B-JAFC}&84.38&60.4&1.61&105.55&79.34&0.88&86.78&56.83&2.1\cr
                    \textit{Mistral-7B-JAFC}&150.18&139.02&0.38&160.62&143.7&0.82&128.47&94.66&9.55\cr
                    \midrule
                    \textit{LLaMA-7B-SelfCheckGPT}&54.08&33.56&0.28&49.99&26.47&0.55&58.67&25.56&4.67\cr
                    \textit{Mistral-7B-SelfCheckGPT}&\textbf{11.43}&\textbf{1.34}&\textbf{0.21}&\textbf{21.72}&\textbf{9.65}&\textbf{0.03}&\textbf{24.17}&\textbf{3.84}&\textbf{0.95}\cr
				\bottomrule
			\end{tabular}
			\caption{Evaluating calibration of various confidence methods. Here, we compare \textit{Just Asking for Calibration (JAFC)}~\cite{tian2023just} and \textit{SelfCheckGPT~\cite{manakul2023selfcheckgpt}}. CL and GL mean calibration loss and grouping loss, respectively. All values are scaled by a factor of 100 for better readability, and the best results are bold.}%
			\label{tab:ori_calibration_results}%
			\vspace{-5pt}
		\end{threeparttable} 
	}
\end{table*}

\paragraph{Methods of Eliciting Confidence.}
We consider two Black-box methods for eliciting confidence scores: \emph{\textit{Just Ask for Calibration}}~\cite{tian2023just} and \emph{\textit{SelfCheckGPT}}~\cite{manakul2023selfcheckgpt}. 
Note that our framework is applicable to other confidence methods as well.

\textit{Just Ask for Calibration (JAFC)} uses dedicated prompts to elicit verbalized probabilities, which can yield better calibrations than the model’s conditional probabilities. We follow the \emph{Verb. 1S top-n} setting to extract numerical probabilities.
It makes the LLM produce $n$ guesses with probabilities, and the answer with the highest
score is selected as the final output.
The prompt used is shown in Appendix~\ref{sec_prompt}.

\textit{SelfCheckGPT} detects hallucinations by comparing the consistency of multiple answers to the same query. We use the version of Natural Language Inference (NLI, also known as Textual Entailment) to compute the confidence score. 
NLI determines
whether a premise entails a hypothesis, and classification labels belong to $\{$\textit{entailment, neutral, contradiction}$\}$ (see, e.g., \cite{tina} for a formal probabilistic definition).
Given a query $q$, we ask an LLM to obtain a main response, which can be regarded as a hypothesis with $m$ sentences $\{r_1, r_2,...,r_m\}$. Then, we use the same query again to ask the LLM $n$ times for obtaining the premise documents $\mathcal{D} = \{d_1, d_2,...,d_n\}$. 
The NLI contradiction score is computed as: 
\begin{equation}
    P(\text{contradict}|r_i, d) = \frac{\exp(z_c)}{\exp(z_e) + \exp(z_c)}
\end{equation}
where $d$ is one premise document, $z_e$ and $z_c$ are the logits of the ``\textit{entailment}''
and ``\textit{contradiction}'' classes, respectively.
This normalization ignores the neutral class and ensures
that the probability is bounded between 0.0 and
1.0, where a higher value means it is more likely to hallucinate. 
The confidence score for each sentence in the main response 
is then defined as: 
\begin{equation}
    S_{\text{SelfCheckGPT}}(r_i) = 1- \frac{1}{m} \sum_{j=1}^{m}P(\text{contradict}|r_i, d_j)
\end{equation}

\paragraph{Evaluation Protocol.}
Since the same entity can have several names (Bill Gates, e.g., is called ``William Henry Gates III''), 
we cannot rely solely on string matching to determine whether the answer of the LLM is correct. 
Therefore, we use an additional NLI model, as follows:
The ground truth in YAGO is converted to a natural sentence, and we judge whether this premise entails the answer by the LLM. 
Moreover, a relation can have several objects per subject. For example, there are two composers for the song \textit{``Rolling in the Deep''}. Therefore, we iterate through all objects in the ground truth and label the LLM answer as correct if it corresponds to any of these objects. 
We manually validated 50 randomly selected samples and all assessments were correct.
We use the DeBERTa~\cite{he2020deberta} model~\footnote{cross-encoder/nli-deberta-v3-large}
fine-tuned on the NLI data set MNLI~\cite{williams2018broad}. 

\paragraph{Metrics.}
Given the observed binary labels $Y$, the true posterior probabilities $Q$, confidence scores $S$ obtained from a model $P(Y)$, and the corresponding average true posterior probabilities $C$,
the divergence of proper scoring rules can be decomposed as~\cite{kull2015novel,perez2022beyond}:
\begin{equation}
\begin{aligned}
\mathbb{E} \left[ f(S, Y) \right] = & \underbrace{\mathbb{E} \left[ f(S, C) \right]}_{\text{Calibration Loss}} \\
& + \underbrace{\mathbb{E} \left[ f(C, Q) \right]}_{\text{Grouping Loss}} + \underbrace{\mathbb{E} \left[ f(Q, Y) \right]}_{\text{Irreducible Loss}}
\end{aligned}
\end{equation}
where $f$ is a function that measures the divergence between the two inputs.
In this work, we consider three metrics: the Brier Score $f^{\text{BS}}(S, Y)$~\cite{brier1950verification}, the Calibration Loss $f^{\text{CL}}(S, C) $, and the Grouping Loss $f^{\text{GL}}(C, Q) $ ~\cite{kull2015novel,perez2022beyond}. (1) The \emph{Brier score} is the squared error between the observed binary labels $Y$ --denoting correct/incorrect answers-- and the associated confidence scores $S$. The appealing property of the Brier score is that it is minimum when $S = P(y)$.
(2) \emph{Calibration Loss (CL)} measures the error rate (average observed $y$) for a given confidence score S: $\mathbb{E}[y | S = s]$; a calibration plot, as in Figure~\ref{fig:birth_date_a} plots this value for different values of $c$. When the confidence score $S$ equals the probability $P(y)$, the calibration plot is on the diagonal, and the calibration error is zero. However, the converse is not true: a calibration error can be zero and yet the confidence score differs from from the probability $P(y)$. The reason for this difference is that within the observations with a predicted confidence score of $S$, some have an actual probability above $C$ while others below: errors compensate \cite{perez2022beyond}.
(3) \emph{Grouping Loss (GL)} 
is the loss due to many instances being
grouped under the same estimate $S$ while having different true posterior
probabilities $Q$~\cite{kull2015novel}.
We reuse the method by \citet{perez2022beyond} to estimate the lower bound of the grouping loss by looking at the dispersion in the error rate on sub-groups of observations for a given score $S$.

\subsection{Evaluating the Calibration of LLMs}~\label{sec:eval_calibration}


The results of our evaluation are shown in Table~\ref{tab:ori_calibration_results}. 
We can see that \textit{Mistral-7B-SelfCheckGPT} performs the best across all tasks, indicating better calibration performance compared to other configurations. 
Notably, \textit{SelfCheckGPT} consistently outperforms \textit{JAFC}, highlighting the inadequacy of relying solely on prompt-based methods. 
Although the three metrics for \textit{Mistral-7B-SelfCheckGPT} appear relatively low, suggesting seemingly acceptable confidence scores, it is crucial to note the existence of sub-groups that are far from well-calibrated.
For example, sub-group analysis within the birth date subset, based on entity popularity and nationality, reveals the model's poor performance for groups with infrequent persons (Figure~\ref{fig:birth_date_b}) and Asian nationalities (Figure~\ref{fig:birth_date_c}).   
This phenomenon confirms that a model may have a low calibration error but there might be sub-groups whose confidence scores deviate dramatically from the true probabilities.

\subsection{Identifying the Grouping Loss in LLMs}~\label{sec:identify_gl}

\begin{figure}[t]%
	\centering
    \subfloat[\centering \textit{LLaMA-7B-SCGPT}]{{\includegraphics[width=0.24\textwidth]{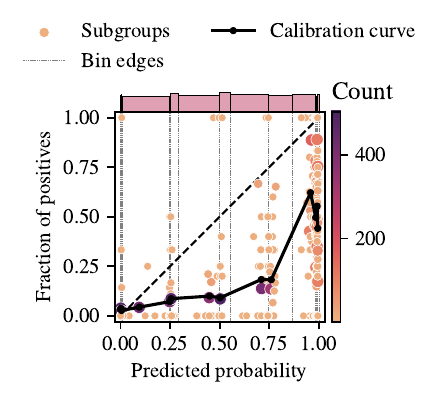} }}%
    \subfloat[\centering \textit{Mistral-7B-SCGPT}]{{\includegraphics[width=0.24\textwidth]{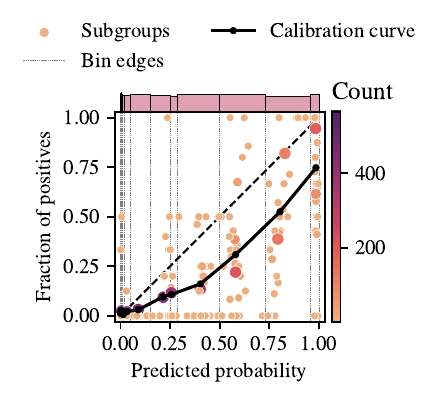} }\label{fig:latent_two_mistral}}
 	\caption{Grouping diagrams of latent sub-groups. These groups are created from the leaves of a  decision tree. SCGPT is an abbreviation for SelfCheckGPT.}
	\label{fig:latent_two}%
\end{figure}

Table~\ref{tab:ori_calibration_results} has already shown the concrete values of grouping loss for different methods.
However, it is not very clear where the grouping loss originated. 
To answer this question, we visualize the behaviors of sub-groups in each method. 

\paragraph{Sub-group Definitions.}~\label{para:subgroup}
We study two types of sub-groups: \emph{user-defined} and \emph{latent} sub-groups.
For \emph{user-defined groups}, we look at explicit features such as popularity, nationality, and gender. We split all samples into different groups based on the entity feature of queries.
User-defined groups may not be adapted to the actual sources of heterogeneity in the confidence score. Therefore, we also use optimized groups that give a tight bound on the grouping loss.
For these \emph{latent groups}, we follow \citet{perez2022beyond} to employ a decision tree, using a loss related to the squared loss for the Brier score on labels ($Y$). This tree defines sub-groups that minimize the loss on a given set of predicted confidence scores. To prevent overfitting, a train-test split is applied: a feature space partition is created using the leaves of the tree fitted on one portion. The input for the decision tree is the embedding of the top layer of an LLM for a particular query. In this way, samples with similar over-confidence / under-confidence can be grouped together. For example, queries featuring well-known entities may be grouped together because an LLM excels at handling them, while queries involving long-tail entities could form a separate group. In practice, groups are defined over multiple different features of queries and are thus much more subtle.

\paragraph{Grouping Diagrams.} In a binary setting, calibration curves display the calibrated scores versus the confidence scores of the positive class, as depicted in Figure~\ref{fig:birth_date_a}. To visualize the heterogeneity among distinct sub-groups within a specific bin, we enrich this representation by including estimated scores for each sub-group, indicating the fraction of positives in each. As shown in Figure~\ref{fig:latent_two}, a larger separation among sub-groups means that the grouping loss is more significant. In this diagram, we use quantile binning with 15 bins. 

Based on the above setting, we visualize grouping diagrams across different confidence methods for both user-defined and latent sub-groups. We aggregate the scores of three relations in this experiment. The results of latent groups are shown in Figure~\ref{fig:latent_two}, while the results of user-defined groups are shown in Figure~\ref{fig:physical} in the appendix.

\begin{table*}[bt] 
	\centering
        \scriptsize
	\setlength{\tabcolsep}{3.95mm}{
		\begin{threeparttable} 
			\begin{tabular}{c|ccc|ccc|ccc}  
				\toprule
				\textbf{Method}
				&\multicolumn{3}{c|}{\textbf{\underline{\texttt{Birth\_Date}}}}&\multicolumn{3}{c|}{\textbf{\underline{\texttt{Founder}}}}&\multicolumn{3}{c}{\textbf{\underline{\texttt{Composer}}}}\cr
                \midrule
			&Brier~\textcolor{brandeisblue}{$\downarrow$}&CL~\textcolor{brandeisblue}{$\downarrow$}&GL~\textcolor{brandeisblue}{$\downarrow$}&Brier~\textcolor{brandeisblue}{$\downarrow$}&CL~\textcolor{brandeisblue}{$\downarrow$}&GL~\textcolor{brandeisblue}{$\downarrow$}&Brier~\textcolor{brandeisblue}{$\downarrow$}&CL~\textcolor{brandeisblue}{$\downarrow$}&GL~\textcolor{brandeisblue}{$\downarrow$}\cr
            \midrule
                \multicolumn{10}{c}{\cellcolor{gray!15} \textit{\footnotesize LLaMA-7B-JAFC}}\cr
                \midrule
                Before &84.38&60.4&1.61&105.55&79.34&0.88&86.78&56.83&2.1\cr
                Calibration &23.79&0.02&1.52&26.39&0.05&0.89&30.06&0.2&2.1\cr
                Ours &\colorbox{blue!20}{22.24}&\colorbox{red!20}{0.03}&\colorbox{blue!20}{0.89}&\colorbox{blue!20}{26.12}&\colorbox{red!20}{0.14}&\colorbox{blue!20}{0.44}&\colorbox{blue!20}{28.81}&\colorbox{red!20}{0.37}&\colorbox{blue!20}{1.36}\cr
                 \midrule
                \multicolumn{10}{c}{\cellcolor{gray!15} \textit{\footnotesize Mistral-7B-JAFC}}\cr
                \midrule
                Before &150.18&139.02&0.38&160.62&143.7&0.82&128.47&94.66&9.55\cr
Calibration &11.14&0.01&0.36&17.24&0.14&0.85&34.1&0.04&9.13\cr
Ours &\colorbox{blue!20}{10.95}&\colorbox{red!20}{0.05}&\colorbox{blue!20}{0.14}&\colorbox{blue!20}{16.97}&\colorbox{red!20}{0.15}&\colorbox{blue!20}{0.34}&\colorbox{blue!20}{26.61}&\colorbox{red!20}{0.17}&\colorbox{blue!20}{0.89}\cr
                \midrule
                \multicolumn{10}{c}{\cellcolor{gray!15} \textit{\footnotesize LLaMA-7B-SelfCheckGPT}}\cr
                \midrule
                    Before &54.08&33.56&0.28&49.99&26.47&0.55&58.67&25.56&4.67\cr
                    Calibration &20.59&0.45&0.76&23.83&0.17&0.74&33.94&0.16&8.83\cr
                    Ours &\colorbox{blue!20}{19.64}&\colorbox{blue!20}{0.24}&\colorbox{blue!20}{0.21}&\colorbox{blue!20}{23.13}&\colorbox{red!20}{0.4}&\colorbox{blue!20}{0.51}&\colorbox{blue!20}{27.06}&\colorbox{red!20}{0.45}&\colorbox{blue!20}{0.93}\cr
                    \midrule
                    \multicolumn{10}{c}{\cellcolor{gray!15} \textit{\footnotesize Mistral-7B-SelfCheckGPT}}\cr
                    \midrule
                    Before &11.43&1.34&0.21&21.72&9.65&0.03&24.17&3.84&0.95\cr
Calibration &10.25&0.05&0.01&12.21&0.14&0.0&20.27&0.18&1.14\cr
Ours &\colorbox{blue!20}{10.21}&\colorbox{red!20}{0.08}&\colorbox{blue!20}{0.0}&\colorbox{blue!20}{12.01}&\colorbox{red!20}{0.15}&\colorbox{blue!20}{0.0}&\colorbox{blue!20}{18.98}&\colorbox{blue!20}{0.13}&\colorbox{blue!20}{0.0}\cr
\midrule
\multicolumn{10}{c}{\cellcolor{gray!15} \textit{\footnotesize LLaMA-13B-SelfCheckGPT}}\cr
\midrule
Before &64.48&33.93&3.01&70.47&40.71&0.23&70.26&32.83&1.34\cr
Calibration &30.96&0.4&4.02&30.22&0.1&1.31&37.36&0.57&1.48\cr
Ours &\colorbox{blue!20}{26.63}&\colorbox{blue!20}{0.33}&\colorbox{blue!20}{0.23}&\colorbox{blue!20}{29.32}&\colorbox{red!20}{0.56}&\colorbox{blue!20}{0.21}&\colorbox{blue!20}{33.78}&\colorbox{red!20}{1.18}&\colorbox{blue!20}{0.58}\cr
\midrule
\multicolumn{10}{c}{\cellcolor{gray!15} \textit{\footnotesize Mixtral-8x7B-SelfCheckGPT}}\cr
\midrule
Before &NA&NA&NA&49.96&27.4&0.1&54.02&23.74&1.27\cr
Calibration &NA&NA&NA&23.82&0.98&0.48&31.42&0.91&0.66\cr
Ours &NA&NA&NA&\colorbox{blue!20}{23.61}&\colorbox{blue!20}{0.61}&\colorbox{blue!20}{0.0}&\colorbox{blue!20}{29.26}&\colorbox{red!20}{1.28}&\colorbox{blue!20}{0.0}\cr
    \bottomrule
    \end{tabular}
    \caption{Comparing methods of after Calibration and our reconfidencing.  Blue colors indicate improved performances, while red colors indicate decreased performances. All values are scaled by a factor of 100 for better readability. Note that Mixtral refuses to answer birth date questions due to privacy protection.}%
			\label{tab:reconfidencing_results}%
			\vspace{-5pt}
		\end{threeparttable} 
	}
\end{table*}

\paragraph{LLMs tend to be overconfident.} 
Ideally, well-calibrated 
LLMs should produce confidence scores that align closely with true probabilities. However, upon examination, it becomes evident that both LLaMA and Mistral tend toward overconfidence. 
Even in the case of \textit{Mistral-7B-SCGPT} (Figure~\ref{fig:latent_two_mistral}), which demonstrates the best performance among other methods, the estimated confidence scores surpass the actual probabilities. For instance, when considering the fraction of true positives at 0.20, the associated confidence score is around 0.50.

\paragraph{The grouping loss is significant. } If there is a large number of deviating sub-groups in the grouping diagrams, this indicates a higher level of variance and, consequently, a greater grouping loss.
Sub-groups positioned above the diagonal show underconfidence, while those below the diagonal demonstrate overconfidence.
Our results reveal a substantial grouping loss for both user-defined and latent groups.
Regarding user-defined groups (Figure~\ref{fig:physical}), we see distinct behaviors among sub-groups based on popularity.  
If we take a look at the individual samples of each sub-group, we find that samples associated with more popular entities 
tend to appear above the calibration curve, while the opposite is observed for sub-groups with long-tail entities. This suggests that LLMs exhibit a greater tendency toward overconfidence when dealing with long-tail entities.

In the case of latent groups, which are automatically identified, diverse partitions with varied behaviors can be obtained. Figure~\ref{fig:latent_two} illustrates a more scattered distribution of sub-groups, including instances of underconfidence not visible through the user-defined groups.

In summary, our analysis indicates a prevalent tendency of overconfidence in LLMs. Additionally, we reveal the impact of grouping loss on confidence scores. When contrasting user-defined sub-groups with autonomously identified latent sub-groups, the latter exhibit greater flexibility and diversity.

\begin{figure*}[t]%
	\centering

\subfloat[\centering Before (\textit{Mistral-7B})]{{\includegraphics[width=0.32\textwidth]{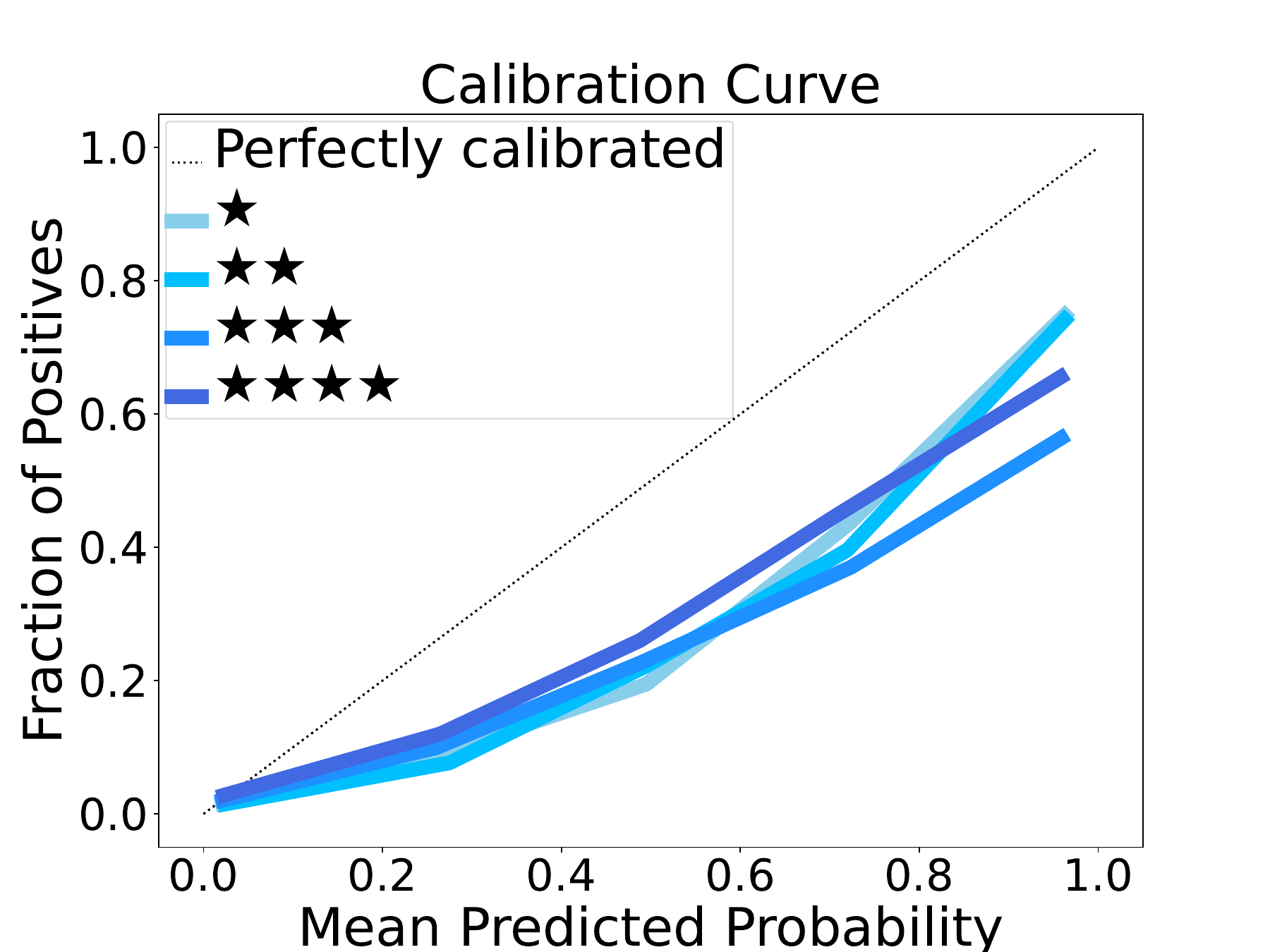} }\label{fig:mistral7b_nli_before}}
	\subfloat[\centering After Calibration (\textit{Mistral-7B})]{{\includegraphics[width=0.32\textwidth]{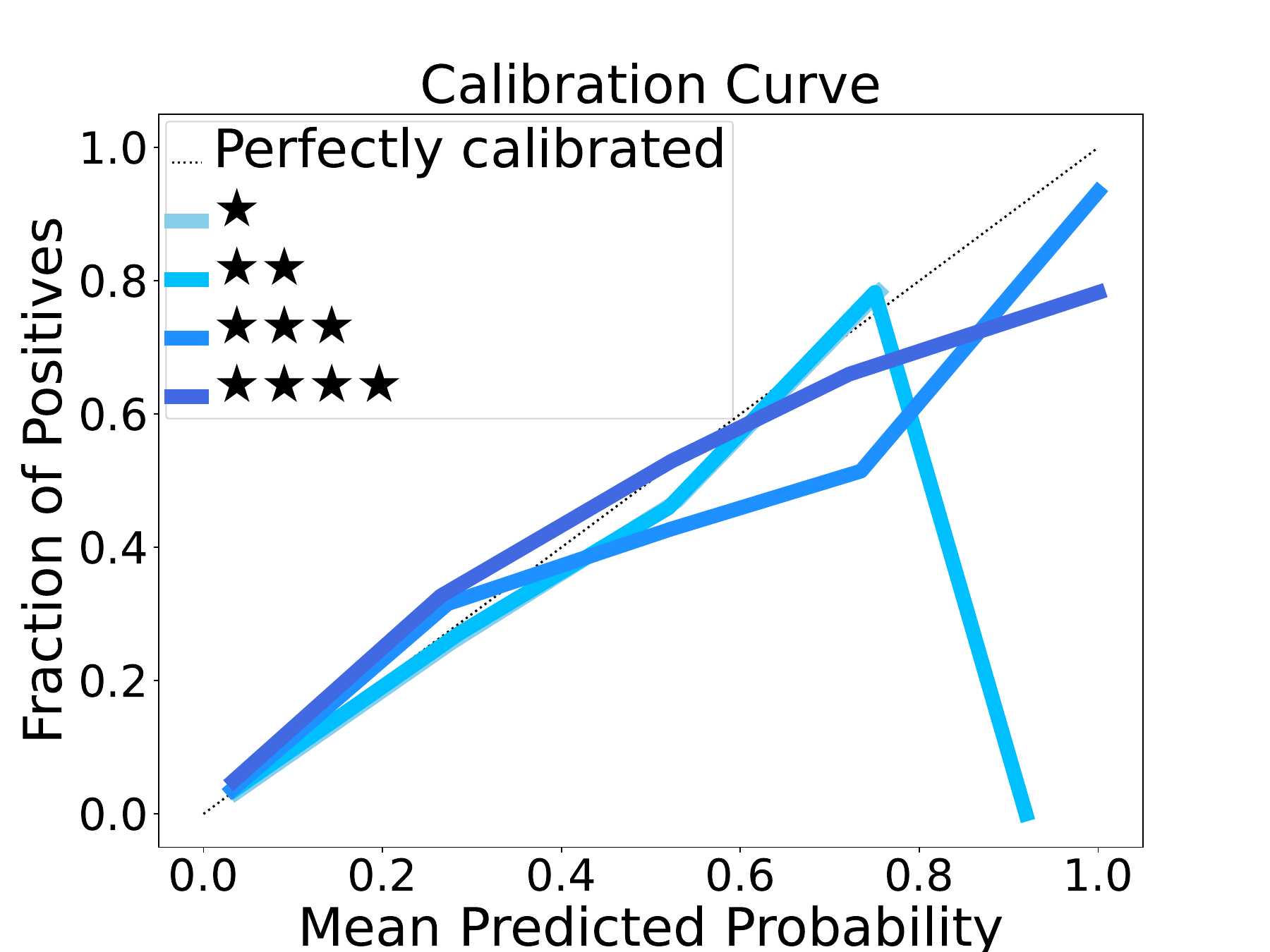} }\label{fig:bmistral7b_nli_recal}}%
 \subfloat[\centering Ours (\textit{Mistral-7B})]{{\includegraphics[width=0.32\textwidth]{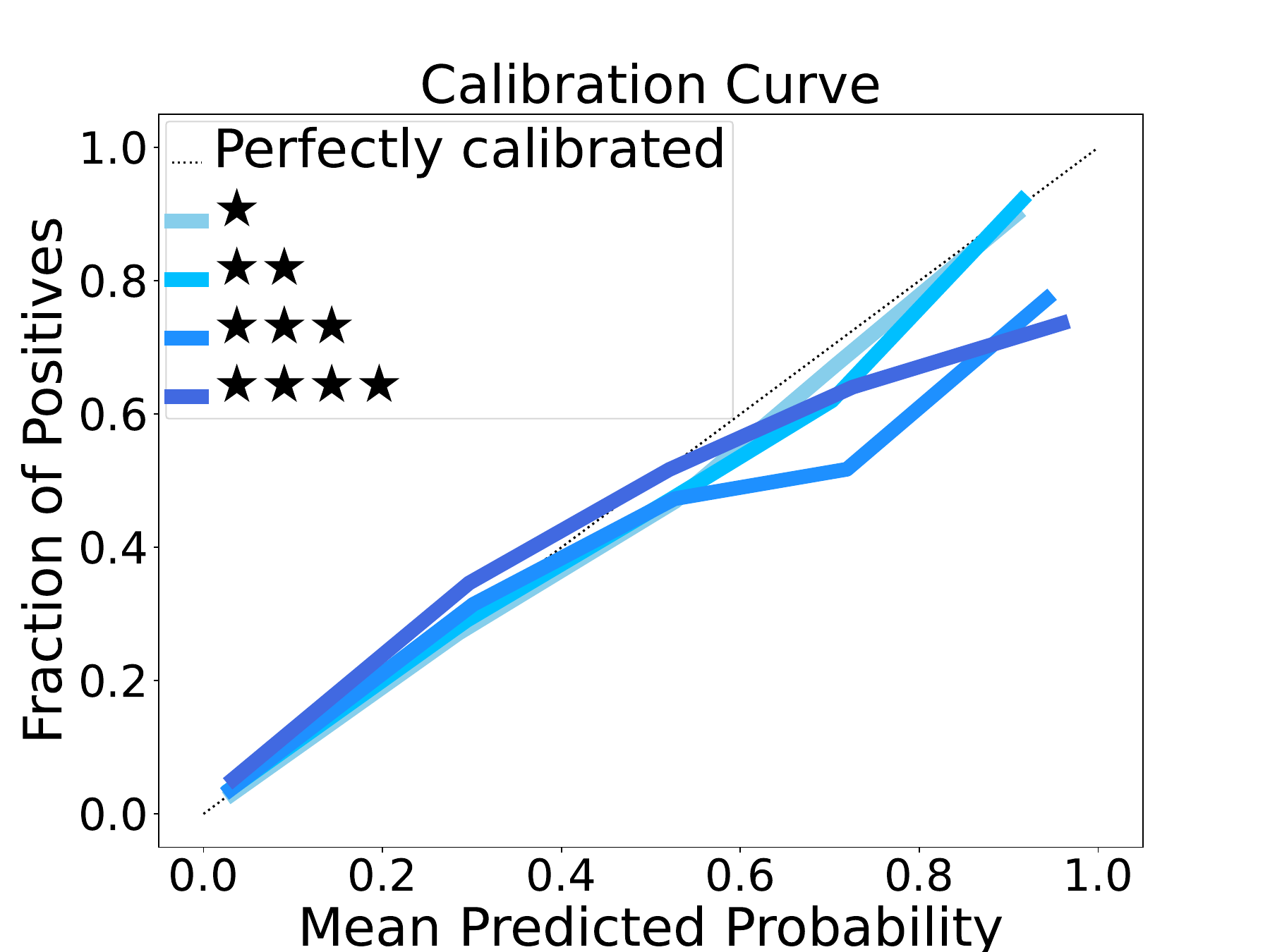} }\label{fig:mistral7b_nli_reconf}}%


 
\caption{Comparing calibrations across different popularity groups for the Mistral-7B. We use merged results of three regions. The confidence method here is SelfCheckGPT. More $\star$ symbols mean a sub-group with more popular samples.}
	\label{fig:before_after_popularity}%
 \vspace{-10pt}
\end{figure*}

\section{Reconfidencing LLMs}

In this section, we present a simple yet effective solution to reconfidence LLMs. The core idea is to calibrate each sub-group separately.

\paragraph{Standard Calibration}  
Following standard calibration procedures, we train a regressor, commonly known as a calibrator, to conduct the calibration of a model \cite{niculescu2005predicting}
This calibrator works by mapping the model's output 
to a refined probability within the interval [0, 1], with the aim of aligning closely with the true probability. Concretely, we train an isotonic regressor using our constructed training and validation sets for calibration purposes \cite{zadrozny2002transforming}.
Subsequently, we apply this trained regressor to calibrate the confidence scores on the test set.

\paragraph{Reconfidencing} 
The standard calibration approaches are marginal: they control average error on confidence and overlook the nuances of sub-groups, where confidence errors can be especially marked.
Inspired by this, we propose a more refined method to calibrate 
LLMs from the sub-group perspective. Adapting \citet{perez2022beyond}, a tree classifier is trained to know how to partition samples (see details in Section~\ref{para:subgroup}). 
We employ a loss function derived from the squared loss for the Brier score on labels ($Y$) to optimize the predicted confidence scores. This decision tree algorithm partitions the data into sub-groups that minimize the specified loss. The tree's input consists of embeddings from the top layer of a LLM for a given query, which can effectively cluster samples exhibiting similar levels of over-confidence or under-confidence.
This, in contrast to user-defined sub-groups,
does not need background knowledge and thus applies to queries that are not matched to the knowledge base. 
Following this step, a distinct isotonic regressor is trained for each identified sub-group.
The final step is to apply this refined method to reconfidence the test set 
The reconfidencing can effectively reduce the grouping loss thus yielding improved calibration results.

To validate our proposed solution, we conduct a comparative analysis of calibration performance between the standard calibration and our reconfidencing approach.  The partition number of the decision tree is eight in this experiment (check Section~\ref{sec:partition_number} to see how we select the leaf number).
Table~\ref{tab:reconfidencing_results} presents the calibration performances of various methods across different relations and LLMs. 
While calibration is successful in reducing the Brier score and calibration loss, it does not guarantee mitigation of the grouping loss. For instance, in the case of \textit{Mistral-7B-SelfCheckGPT} on the composer relation, the calibration significantly improves the Brier score ($24.1 \rightarrow 20.27$) and calibration loss ($3.84 \rightarrow 0.18$). However, it is noteworthy that the grouping loss increases ($0.95 \rightarrow 1.14$). 
Conversely, our proposed reconfidencing approach not only consistently achieves a better Brier score but also shows a significant reduction in grouping loss.
Using the same example, our method attains a lower Brier score ($20.27 \rightarrow 18.98$) and effectively eliminates grouping loss ($1.14 \rightarrow 0.0$) compared to the calibration method.

Since our reconfidencing works on the latent group loss, it does not specifically target the issues shown in the examples of popularity (Figure~\ref{fig:birth_date_b}) and nationality (Figure~\ref{fig:birth_date_c}). 
To answer whether it improves the situation for these user-defined groups, we analyze calibration curves across samples after calibration and reconfidencing.
 The results for popularity and nationality sub-groups are shown in Figure~\ref{fig:before_after_popularity} and Figure~\ref{fig:recal_ours_nationality} respectively.
Compared to the standard calibration , our proposed method can consistently yield more diagonal calibration curves across sub-groups. 

To show the scalability of our method on other relations and other types of groups, we conduct experiments on \texttt{Birth\_Place} and \texttt{LocationCreated}. 
Experimental results confirm again that our model can reduce biased information on gender group (Figure~\ref{fig:gender_group}) and the location relation (Figure~\ref{fig:location_created}).
The same observed improvements can also be extended to different sizes of LLaMA (Figure~\ref{fig:popularity_7b_13b}).

\section{Conclusion}
In this work, we analyzed how trustworthy current methods are when they give confidence scores to LLM answers. 
We create a novel dataset derived from the ground truth within the YAGO knowledge base, providing a framework for evaluating the calibration of confidence scores for different groups. 
Subsequent evaluations of different sizes of LLMs reveal a consistent discrimination towards particular minority groups. 
We show grouping loss in LLMs,  such as those associated with long-tail entities and individuals of Asian origin.
These findings emphasize that we should pay particular attention to minority groups when calibrating LLMs.    
Building upon these insights, we introduce a novel approach for reconfidencing LLMs based on latent sub-groups, resulting in improved calibrations.  
This new approach can mitigate the problem of hallucinations by generating alerts in response to LLM answers. 
Meanwhile, our findings can reduce biased information against groups such as race and gender, which is useful for the fairness of LLMs.

\section*{Limitations}
One limitation of our proposed method is that it targets entity-related questions, and not long-form open-ended tasks, as shown in Section~\ref{sec:open_qa} in the appendix. For example, there is no obvious benefit of our method for this very common question: \textit{``why is the sky blue?''} from the TruthfulQA dataset~\cite{lin2022truthfulqa}.
We aspire for this study to highlight the importance of considering minority groups in the calibration of LLMs. Additionally, we anticipate that future research can build upon our methodology to encompass open-ended generation tasks.

\section*{Acknowledgements}
This work was partially funded by projects NoRDF
(ANR-20-CHIA-0012-01) and LearnI (ANR-20-
CHIA-0026).

\bibliography{anthology,custom}

\begin{thebibliography}{41}
\expandafter\ifx\csname natexlab\endcsname\relax\def\natexlab#1{#1}\fi

\bibitem[{Agrawal et~al.(2023)Agrawal, Mackey, and Kalai}]{agrawal2023language}
Ayush Agrawal, Lester Mackey, and Adam~Tauman Kalai. 2023.
\newblock \href {https://arxiv.org/abs/2305.18248} {Do language models know when they're hallucinating references?}
\newblock \emph{ArXiv preprint}.

\bibitem[{Azaria and Mitchell(2023)}]{azaria2023internal}
Amos Azaria and Tom Mitchell. 2023.
\newblock \href {https://arxiv.org/abs/2304.13734} {The internal state of an llm knows when its lying}.
\newblock \emph{ArXiv preprint}.

\bibitem[{Baan et~al.(2023)Baan, Daheim, Ilia, Ulmer, Li, Fern{\'a}ndez, Plank, Sennrich, Zerva, and Aziz}]{baan2023uncertainty}
Joris Baan, Nico Daheim, Evgenia Ilia, Dennis Ulmer, Haau-Sing Li, Raquel Fern{\'a}ndez, Barbara Plank, Rico Sennrich, Chrysoula Zerva, and Wilker Aziz. 2023.
\newblock \href {https://arxiv.org/abs/2307.15703} {Uncertainty in natural language generation: From theory to applications}.
\newblock \emph{ArXiv preprint}.

\bibitem[{Brier(1950)}]{brier1950verification}
Glenn~W Brier. 1950.
\newblock Verification of forecasts expressed in terms of probability.
\newblock \emph{Monthly weather review}, (1).

\bibitem[{Fu et~al.(2023)Fu, Ng, Jiang, and Liu}]{fu2023gptscore}
Jinlan Fu, See-Kiong Ng, Zhengbao Jiang, and Pengfei Liu. 2023.
\newblock \href {https://arxiv.org/abs/2302.04166} {Gptscore: Evaluate as you desire}.
\newblock \emph{ArXiv preprint}.

\bibitem[{Galárraga et~al.(2015)Galárraga, Teflioudi, Hose, and Suchanek}]{amieplus}
Luis Galárraga, Christina Teflioudi, Katja Hose, and Fabian~M. Suchanek. 2015.
\newblock { Fast Rule Mining in Ontological Knowledge Bases with AMIE+ }.
\newblock In \emph{{VLDBJ}}.

\bibitem[{Gawlikowski et~al.(2021)Gawlikowski, Tassi, Ali, Lee, Humt, Feng, Kruspe, Triebel, Jung, Roscher et~al.}]{gawlikowski2021survey}
Jakob Gawlikowski, Cedrique Rovile~Njieutcheu Tassi, Mohsin Ali, Jongseok Lee, Matthias Humt, Jianxiang Feng, Anna Kruspe, Rudolph Triebel, Peter Jung, Ribana Roscher, et~al. 2021.
\newblock \href {https://arxiv.org/abs/2107.03342} {A survey of uncertainty in deep neural networks}.
\newblock \emph{ArXiv preprint}.

\bibitem[{Gou et~al.(2023)Gou, Shao, Gong, Shen, Yang, Duan, and Chen}]{gou2023critic}
Zhibin Gou, Zhihong Shao, Yeyun Gong, Yelong Shen, Yujiu Yang, Nan Duan, and Weizhu Chen. 2023.
\newblock \href {https://arxiv.org/abs/2305.11738} {Critic: Large language models can self-correct with tool-interactive critiquing}.
\newblock \emph{ArXiv preprint}.

\bibitem[{Guo et~al.(2017)Guo, Pleiss, Sun, and Weinberger}]{guo2017calibration}
Chuan Guo, Geoff Pleiss, Yu~Sun, and Kilian~Q. Weinberger. 2017.
\newblock \href {http://proceedings.mlr.press/v70/guo17a.html} {On calibration of modern neural networks}.
\newblock In \emph{Proc. of ICML}, Proceedings of Machine Learning Research.

\bibitem[{He et~al.(2021)He, Liu, Gao, and Chen}]{he2020deberta}
Pengcheng He, Xiaodong Liu, Jianfeng Gao, and Weizhu Chen. 2021.
\newblock \href {https://openreview.net/forum?id=XPZIaotutsD} {Deberta: decoding-enhanced bert with disentangled attention}.
\newblock In \emph{Proc. of ICLR}.

\bibitem[{Helwe et~al.(2022)Helwe, Coumes, Clavel, and Suchanek}]{tina}
Chadi Helwe, Simon Coumes, Chloé Clavel, and Fabian~M. Suchanek. 2022.
\newblock {TINA: Textual Inference with Negation Augmentation}.
\newblock In \emph{{EMNLP Find.}}

\bibitem[{Hendrycks et~al.(2021)Hendrycks, Burns, Basart, Zou, Mazeika, Song, and Steinhardt}]{hendrycks2020measuring}
Dan Hendrycks, Collin Burns, Steven Basart, Andy Zou, Mantas Mazeika, Dawn Song, and Jacob Steinhardt. 2021.
\newblock \href {https://openreview.net/forum?id=d7KBjmI3GmQ} {Measuring massive multitask language understanding}.
\newblock In \emph{Proc. of ICLR}.

\bibitem[{Huang et~al.(2023)Huang, Yu, Ma, Zhong, Feng, Wang, Chen, Peng, Feng, Qin et~al.}]{huang2023survey}
Lei Huang, Weijiang Yu, Weitao Ma, Weihong Zhong, Zhangyin Feng, Haotian Wang, Qianglong Chen, Weihua Peng, Xiaocheng Feng, Bing Qin, et~al. 2023.
\newblock \href {https://arxiv.org/abs/2311.05232} {A survey on hallucination in large language models: Principles, taxonomy, challenges, and open questions}.
\newblock \emph{ArXiv preprint}.

\bibitem[{Ji et~al.(2023)Ji, Lee, Frieske, Yu, Su, Xu, Ishii, Bang, Madotto, and Fung}]{ji2023survey}
Ziwei Ji, Nayeon Lee, Rita Frieske, Tiezheng Yu, Dan Su, Yan Xu, Etsuko Ishii, Ye~Jin Bang, Andrea Madotto, and Pascale Fung. 2023.
\newblock Survey of hallucination in natural language generation.
\newblock \emph{ACM Computing Surveys}, (12).

\bibitem[{Jiang et~al.(2024)Jiang, Sablayrolles, Roux, Mensch, Savary, Bamford, Chaplot, Casas, Hanna, Bressand et~al.}]{jiang2024mixtral}
Albert~Q Jiang, Alexandre Sablayrolles, Antoine Roux, Arthur Mensch, Blanche Savary, Chris Bamford, Devendra~Singh Chaplot, Diego de~las Casas, Emma~Bou Hanna, Florian Bressand, et~al. 2024.
\newblock Mixtral of experts.
\newblock \emph{arXiv preprint arXiv:2401.04088}.

\bibitem[{Jiang et~al.(2021)Jiang, Araki, Ding, and Neubig}]{jiang2021can}
Zhengbao Jiang, Jun Araki, Haibo Ding, and Graham Neubig. 2021.
\newblock \href {https://aclanthology.org/2021.tacl-1.57} {How can we know when language models know? on the calibration of language models for question answering}.
\newblock \emph{Transactions of the Association for Computational Linguistics}.

\bibitem[{Joshi et~al.(2017)Joshi, Choi, Weld, and Zettlemoyer}]{joshi2017triviaqa}
Mandar Joshi, Eunsol Choi, Daniel Weld, and Luke Zettlemoyer. 2017.
\newblock \href {https://aclanthology.org/P17-1147} {{T}rivia{QA}: A large scale distantly supervised challenge dataset for reading comprehension}.
\newblock In \emph{Proc. of ACL}.

\bibitem[{Kadavath et~al.(2022)Kadavath, Conerly, Askell, Henighan, Drain, Perez, Schiefer, Hatfield-Dodds, DasSarma, Tran-Johnson et~al.}]{kadavath2022language}
Saurav Kadavath, Tom Conerly, Amanda Askell, Tom Henighan, Dawn Drain, Ethan Perez, Nicholas Schiefer, Zac Hatfield-Dodds, Nova DasSarma, Eli Tran-Johnson, et~al. 2022.
\newblock \href {https://arxiv.org/abs/2207.05221} {Language models (mostly) know what they know}.
\newblock \emph{ArXiv preprint}.

\bibitem[{Kuhn et~al.(2022)Kuhn, Gal, and Farquhar}]{kuhn2022semantic}
Lorenz Kuhn, Yarin Gal, and Sebastian Farquhar. 2022.
\newblock Semantic uncertainty: Linguistic invariances for uncertainty estimation in natural language generation.
\newblock In \emph{NeurIPS ML Safety Workshop}.

\bibitem[{Kull and Flach(2015)}]{kull2015novel}
Meelis Kull and Peter Flach. 2015.
\newblock Novel decompositions of proper scoring rules for classification: Score adjustment as precursor to calibration.
\newblock In \emph{Machine Learning and Knowledge Discovery in Databases: European Conference, ECML PKDD 2015, Porto, Portugal, September 7-11, 2015, Proceedings, Part I 15}. Springer.

\bibitem[{Lin et~al.(2022{\natexlab{a}})Lin, Hilton, and Evans}]{lin2022teaching}
Stephanie Lin, Jacob Hilton, and Owain Evans. 2022{\natexlab{a}}.
\newblock Teaching models to express their uncertainty in words.
\newblock \emph{Transactions on Machine Learning Research}.

\bibitem[{Lin et~al.(2022{\natexlab{b}})Lin, Hilton, and Evans}]{lin2022truthfulqa}
Stephanie Lin, Jacob Hilton, and Owain Evans. 2022{\natexlab{b}}.
\newblock Truthfulqa: Measuring how models mimic human falsehoods.
\newblock In \emph{Proceedings of the 60th Annual Meeting of the Association for Computational Linguistics (Volume 1: Long Papers)}, pages 3214--3252.

\bibitem[{Lin et~al.(2023)Lin, Trivedi, and Sun}]{lin2023generating}
Zhen Lin, Shubhendu Trivedi, and Jimeng Sun. 2023.
\newblock \href {https://arxiv.org/abs/2305.19187} {Generating with confidence: Uncertainty quantification for black-box large language models}.
\newblock \emph{ArXiv preprint}.

\bibitem[{Manakul et~al.(2023)Manakul, Liusie, and Gales}]{manakul2023selfcheckgpt}
Potsawee Manakul, Adian Liusie, and Mark~JF Gales. 2023.
\newblock \href {https://arxiv.org/abs/2303.08896} {Selfcheckgpt: Zero-resource black-box hallucination detection for generative large language models}.
\newblock \emph{ArXiv preprint}.

\bibitem[{Mielke et~al.(2022)Mielke, Szlam, Dinan, and Boureau}]{mielke2022reducing}
Sabrina~J. Mielke, Arthur Szlam, Emily Dinan, and Y-Lan Boureau. 2022.
\newblock \href {https://aclanthology.org/2022.tacl-1.50} {Reducing conversational agents{'} overconfidence through linguistic calibration}.
\newblock \emph{Transactions of the Association for Computational Linguistics}.

\bibitem[{MistralAI(2023)}]{jiang2023mistral}
MistralAI. 2023.
\newblock \href {https://arxiv.org/abs/2310.06825} {Mistral 7b}.
\newblock \emph{ArXiv preprint}.

\bibitem[{Niculescu-Mizil and Caruana(2005)}]{niculescu2005predicting}
Alexandru Niculescu-Mizil and Rich Caruana. 2005.
\newblock Predicting good probabilities with supervised learning.
\newblock In \emph{Proceedings of the 22nd international conference on Machine learning}, pages 625--632.

\bibitem[{OpenAI(2022)}]{chatgpt}
OpenAI. 2022.
\newblock Introducing chatgpt.
\newblock \url{https://openai.com/blog/chatgpt}.

\bibitem[{Ouyang et~al.(2022)Ouyang, Wu, Jiang, Almeida, Wainwright, Mishkin, Zhang, Agarwal, Slama, Ray et~al.}]{ouyang2022training}
Long Ouyang, Jeffrey Wu, Xu~Jiang, Diogo Almeida, Carroll Wainwright, Pamela Mishkin, Chong Zhang, Sandhini Agarwal, Katarina Slama, Alex Ray, et~al. 2022.
\newblock Training language models to follow instructions with human feedback.
\newblock \emph{Advances in Neural Information Processing Systems}.

\bibitem[{Park and Caragea(2022)}]{park2022calibration}
Seo~Yeon Park and Cornelia Caragea. 2022.
\newblock \href {https://aclanthology.org/2022.acl-long.368} {On the calibration of pre-trained language models using mixup guided by area under the margin and saliency}.
\newblock In \emph{Proc. of ACL}.

\bibitem[{Perez-Lebel et~al.(2023)Perez-Lebel, Le~Morvan, and Varoquaux}]{perez2022beyond}
Alexandre Perez-Lebel, Marine Le~Morvan, and Gael Varoquaux. 2023.
\newblock Beyond calibration: estimating the grouping loss of modern neural networks.
\newblock In \emph{The Eleventh International Conference on Learning Representations}.

\bibitem[{Suchanek et~al.(2024)Suchanek, Alam, Bonald, Chen, Paris, and Soria}]{suchanek2023integrating}
Fabian~M. Suchanek, Mehwish Alam, Thomas Bonald, Lihu Chen, Pierre-Henri Paris, and Jules Soria. 2024.
\newblock {YAGO 4.5: A Large and Clean Knowledge Base with a Rich Taxonomy}.

\bibitem[{Tian et~al.(2023)Tian, Mitchell, Zhou, Sharma, Rafailov, Yao, Finn, and Manning}]{tian2023just}
Katherine Tian, Eric Mitchell, Allan Zhou, Archit Sharma, Rafael Rafailov, Huaxiu Yao, Chelsea Finn, and Christopher~D Manning. 2023.
\newblock \href {https://arxiv.org/abs/2305.14975} {Just ask for calibration: Strategies for eliciting calibrated confidence scores from language models fine-tuned with human feedback}.
\newblock \emph{ArXiv preprint}.

\bibitem[{Touvron et~al.(2023)Touvron, Lavril, Izacard, Martinet, Lachaux, Lacroix, Rozi{\`e}re, Goyal, Hambro, Azhar et~al.}]{touvron2023llama}
Hugo Touvron, Thibaut Lavril, Gautier Izacard, Xavier Martinet, Marie-Anne Lachaux, Timoth{\'e}e Lacroix, Baptiste Rozi{\`e}re, Naman Goyal, Eric Hambro, Faisal Azhar, et~al. 2023.
\newblock \href {https://arxiv.org/abs/2302.13971} {Llama: Open and efficient foundation language models}.
\newblock \emph{ArXiv preprint}.

\bibitem[{Welbl et~al.(2017)Welbl, Liu, and Gardner}]{welbl2017crowdsourcing}
Johannes Welbl, Nelson~F Liu, and Matt Gardner. 2017.
\newblock Crowdsourcing multiple choice science questions.
\newblock In \emph{Proceedings of the 3rd Workshop on Noisy User-generated Text}, pages 94--106.

\bibitem[{Williams et~al.(2018)Williams, Nangia, and Bowman}]{williams2018broad}
Adina Williams, Nikita Nangia, and Samuel Bowman. 2018.
\newblock \href {https://aclanthology.org/N18-1101} {A broad-coverage challenge corpus for sentence understanding through inference}.
\newblock In \emph{Proc. of NAACL-HLT}.

\bibitem[{Xiao et~al.(2022)Xiao, Liang, Bhatt, Neiswanger, Salakhutdinov, and Morency}]{xiao2022uncertainty}
Yuxin Xiao, Paul~Pu Liang, Umang Bhatt, Willie Neiswanger, Ruslan Salakhutdinov, and Louis-Philippe Morency. 2022.
\newblock \href {https://aclanthology.org/2022.findings-emnlp.538} {Uncertainty quantification with pre-trained language models: A large-scale empirical analysis}.
\newblock In \emph{Findings of the Association for Computational Linguistics: EMNLP 2022}.

\bibitem[{Xiong et~al.(2023)Xiong, Hu, Lu, Li, Fu, He, and Hooi}]{xiong2023can}
Miao Xiong, Zhiyuan Hu, Xinyang Lu, Yifei Li, Jie Fu, Junxian He, and Bryan Hooi. 2023.
\newblock \href {https://arxiv.org/abs/2306.13063} {Can llms express their uncertainty? an empirical evaluation of confidence elicitation in llms}.
\newblock \emph{ArXiv preprint}.

\bibitem[{Zadrozny and Elkan(2002)}]{zadrozny2002transforming}
Bianca Zadrozny and Charles Elkan. 2002.
\newblock Transforming classifier scores into accurate multiclass probability estimates.
\newblock In \emph{Proceedings of the eighth ACM SIGKDD international conference on Knowledge discovery and data mining}, pages 694--699.

\bibitem[{Zhang et~al.(2023)Zhang, Li, Cui, Cai, Liu, Fu, Huang, Zhao, Zhang, Chen et~al.}]{zhang2023siren}
Yue Zhang, Yafu Li, Leyang Cui, Deng Cai, Lemao Liu, Tingchen Fu, Xinting Huang, Enbo Zhao, Yu~Zhang, Yulong Chen, et~al. 2023.
\newblock \href {https://arxiv.org/abs/2309.01219} {Siren's song in the ai ocean: A survey on hallucination in large language models}.
\newblock \emph{ArXiv preprint}.

\bibitem[{Zhou et~al.(2023)Zhou, Jurafsky, and Hashimoto}]{zhou2023navigating}
Kaitlyn Zhou, Dan Jurafsky, and Tatsunori Hashimoto. 2023.
\newblock \href {https://arxiv.org/abs/2302.13439} {Navigating the grey area: Expressions of overconfidence and uncertainty in language models}.
\newblock \emph{ArXiv preprint}.

\end{thebibliography}
\bibliographystyle{acl_natbib}

\newpage
\appendix

\setcounter{table}{0}   
\setcounter{figure}{0}
\renewcommand{\thetable}{A\arabic{table}}
\renewcommand{\thefigure}{A\arabic{figure}}
\setcounter{equation}{0}
\setcounter{subsection}{0}
\renewcommand{\theequation}{A.\arabic{equation}}

\section{Appendix}

\subsection{Prompts}\label{sec_prompt}
The prompt used for \emph{\textit{SelfCheckGPT}} to elicit confidence scores~\cite{manakul2023selfcheckgpt} is shown below: 
\fbox{\begin{minipage}{18.6em}
\textit{Provide your best guess and the probability that it is correct (0.0 to 1.0) for
the following question. Give ONLY the guess and probability, no other words or
explanation. For example:\textbackslash n\textbackslash n Guess: $<$most likely guess, as short as possible; not
a complete sentence, just the guess!$>$\textbackslash n Probability: $<$the probability between 0.0
and 1.0 that your guess is correct, without any extra commentary whatsoever; just
the probability!$>$\textbackslash n\textbackslash n The question is: \$$\{\text{THE\_QUESTION\}}$
}
\end{minipage}}

\begin{figure*}[t]%
	\centering
	\subfloat[\centering \textit{LLaMA-7B-JAFC}]{{\includegraphics[width=0.24\textwidth]{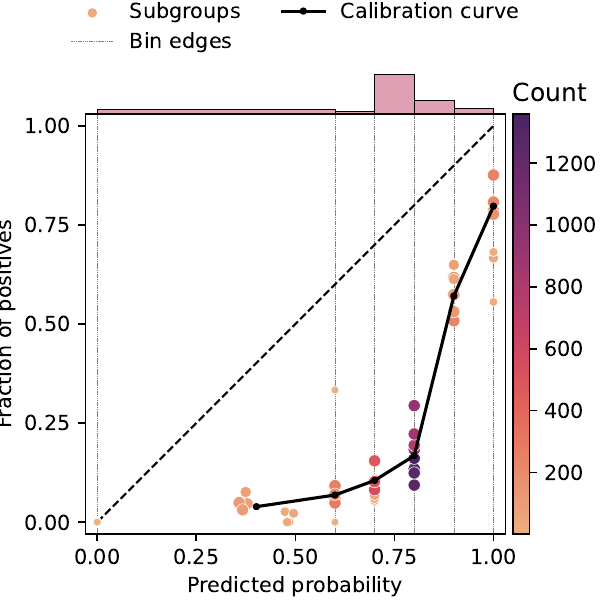} }}%
	\subfloat[\centering \textit{Mistral-7B-JAFC}]{{\includegraphics[width=0.24\textwidth]{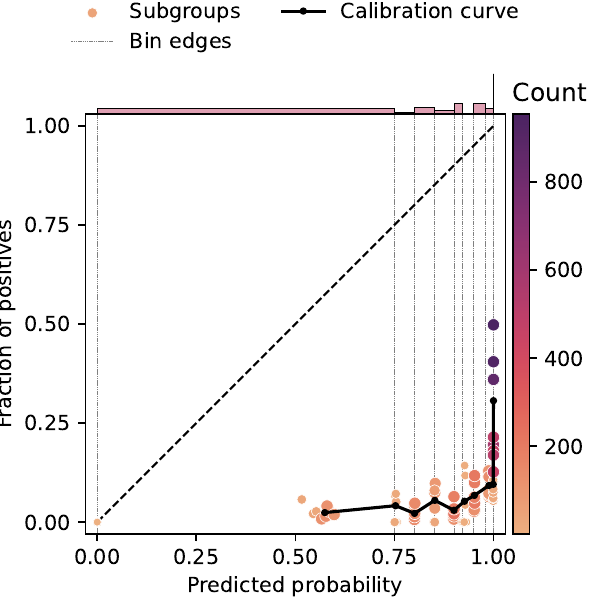} }}%
 \subfloat[\centering \textit{LLaMA-7B-SCGPT}]{{\includegraphics[width=0.24\textwidth]{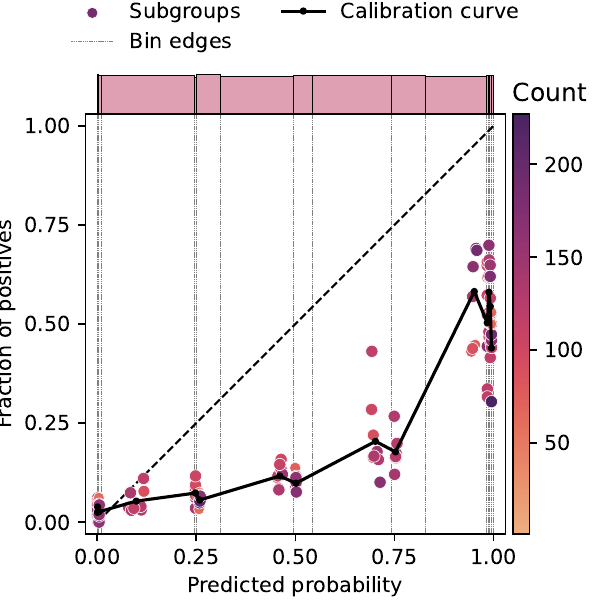} }}%
 \subfloat[\centering \textit{Mistral-7B-SCGPT}]{{\includegraphics[width=0.24\textwidth]{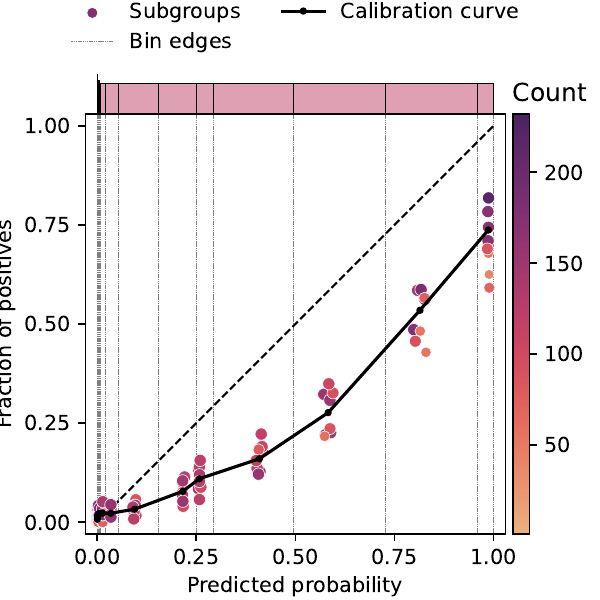} }\label{fig:physical_mistral_nli_S_before}}%
	\caption{Grouping diagrams of user-defined sub-groups. We divide each bin into eight groups by the popularity of entities. SCGPT is an abbreviation for SelfCheckGPT.}
	\label{fig:physical}%
\end{figure*}
\begin{figure*}[t]%
	\centering
	\subfloat[\centering \textit{LLaMA-7B-JAFC}]{{\includegraphics[width=0.24\textwidth]{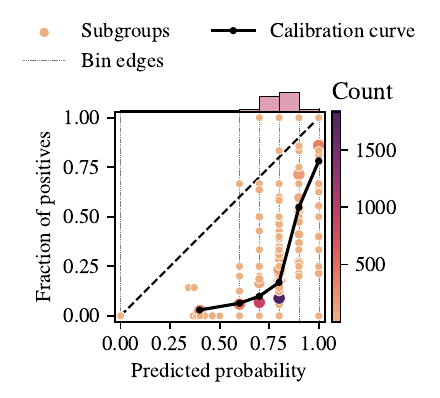} }}%
	\subfloat[\centering \textit{Mistral-7B-JAFC}]{{\includegraphics[width=0.24\textwidth]{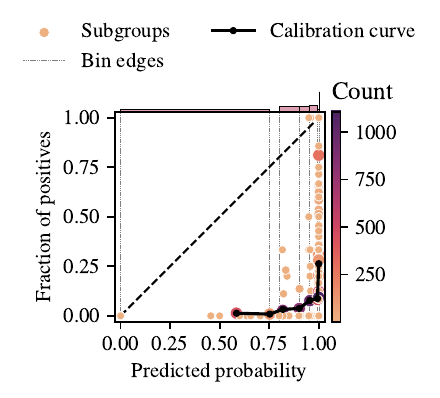} }}%
    \subfloat[\centering \textit{LLaMA-7B-SCGPT}]{{\includegraphics[width=0.24\textwidth]{figures/latent_llama7b_nli.pdf} }}%
    \subfloat[\centering \textit{Mistral-7B-SCGPT}]{{\includegraphics[width=0.24\textwidth]{figures/latent_mistral7b_nli.pdf} }}%
 	\caption{Grouping diagrams of latent sub-groups. These groups are created from the leaves of a  decision tree. SCGPT is an abbreviation for SelfCheckGPT.}
	\label{fig:latent}%
\end{figure*}

\subsection{Reconfidencing Sub-groups}
In this section, we conduct a comparative analysis of the performance between calibration and our proposed reconfidencing. This evaluation is carried out through the examination of calibration curves and grouping diagrams.

\begin{figure*}[bt]%
	\centering
	\subfloat[\centering Before (\textit{LLaMA})]{{\includegraphics[width=0.32\textwidth]{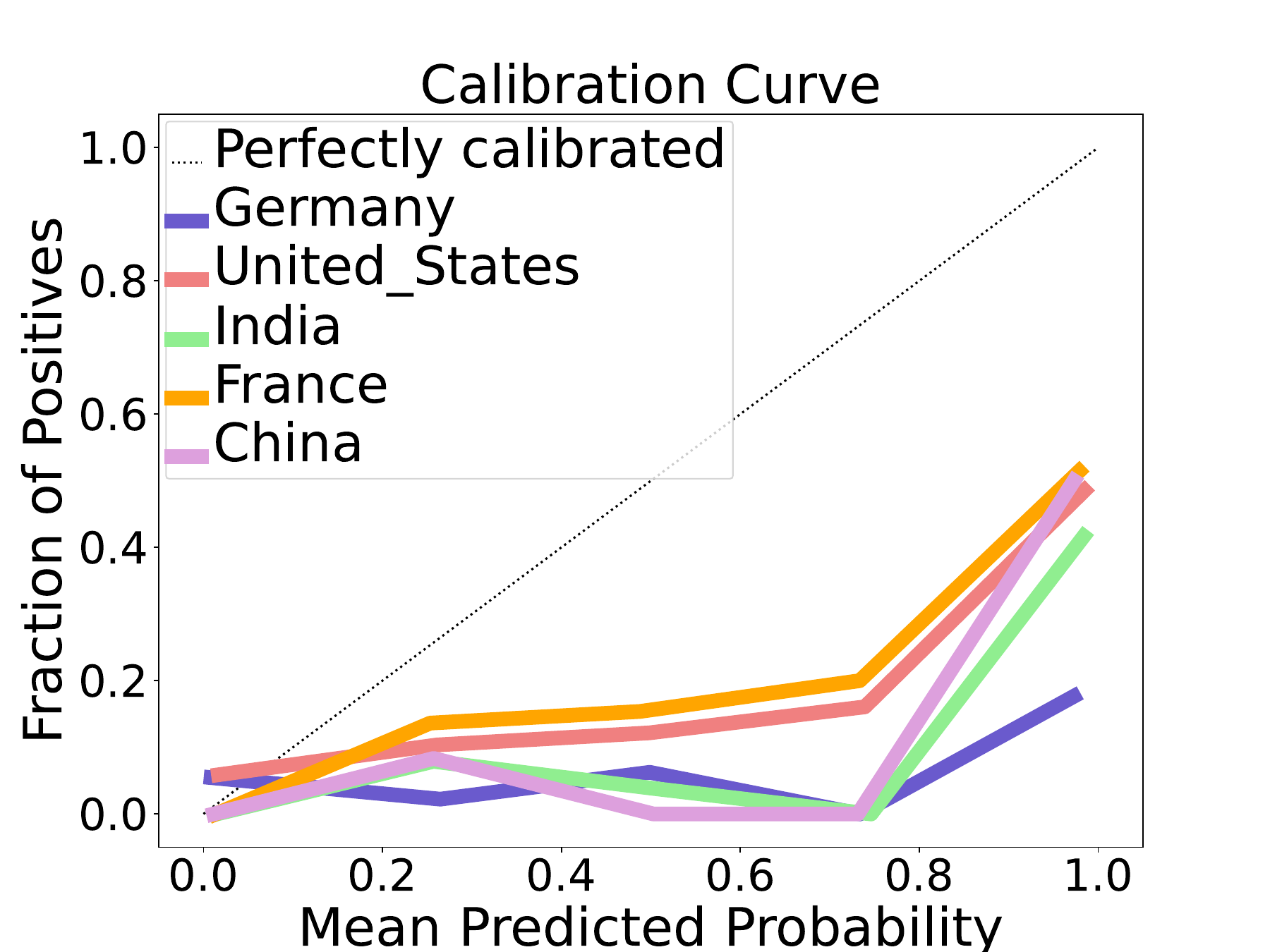} }~\label{fig:nation_llama7b_nli_before}}
	\subfloat[\centering After Calibration (\textit{LLaMA})]{{\includegraphics[width=0.32\textwidth]{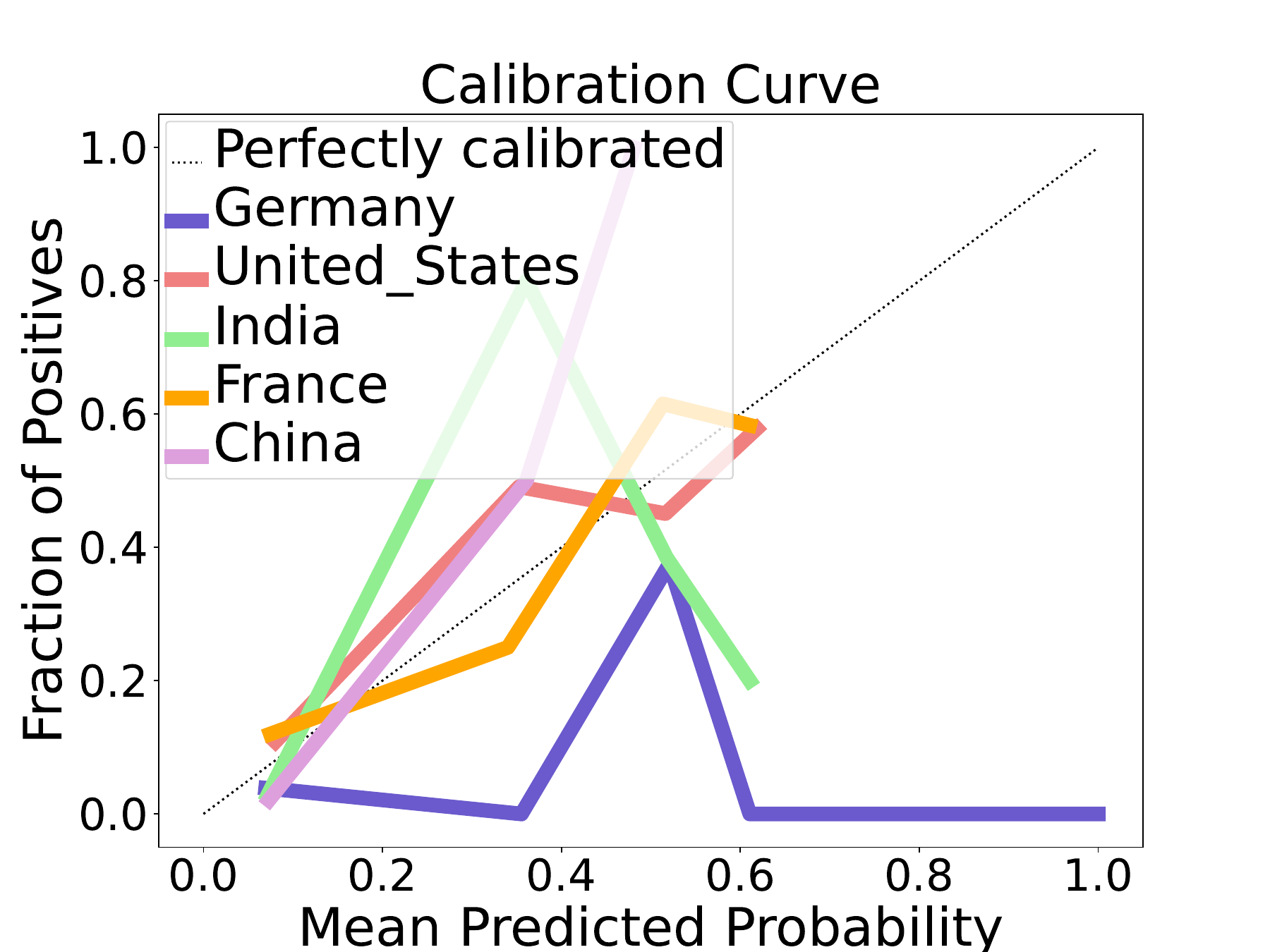} }~\label{fig:nation_llama7b_nli_recal}}%
 \subfloat[\centering Ours (\textit{LLaMA})]{{\includegraphics[width=0.32\textwidth]{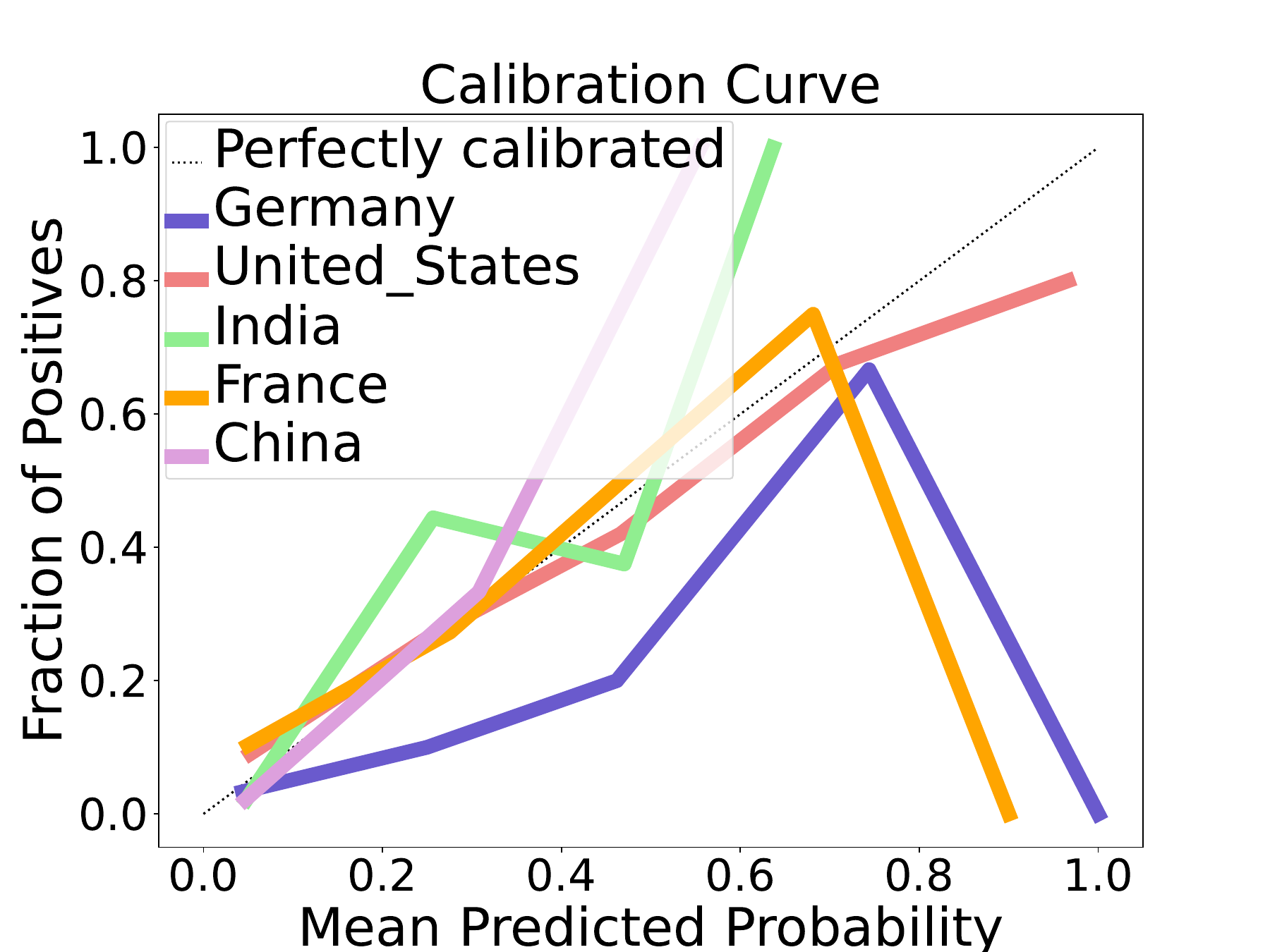} }~\label{fig:nation_llama7b_nli_reconf}}%

    \bigskip
\subfloat[\centering Before (\textit{Mistral})]{{\includegraphics[width=0.32\textwidth]{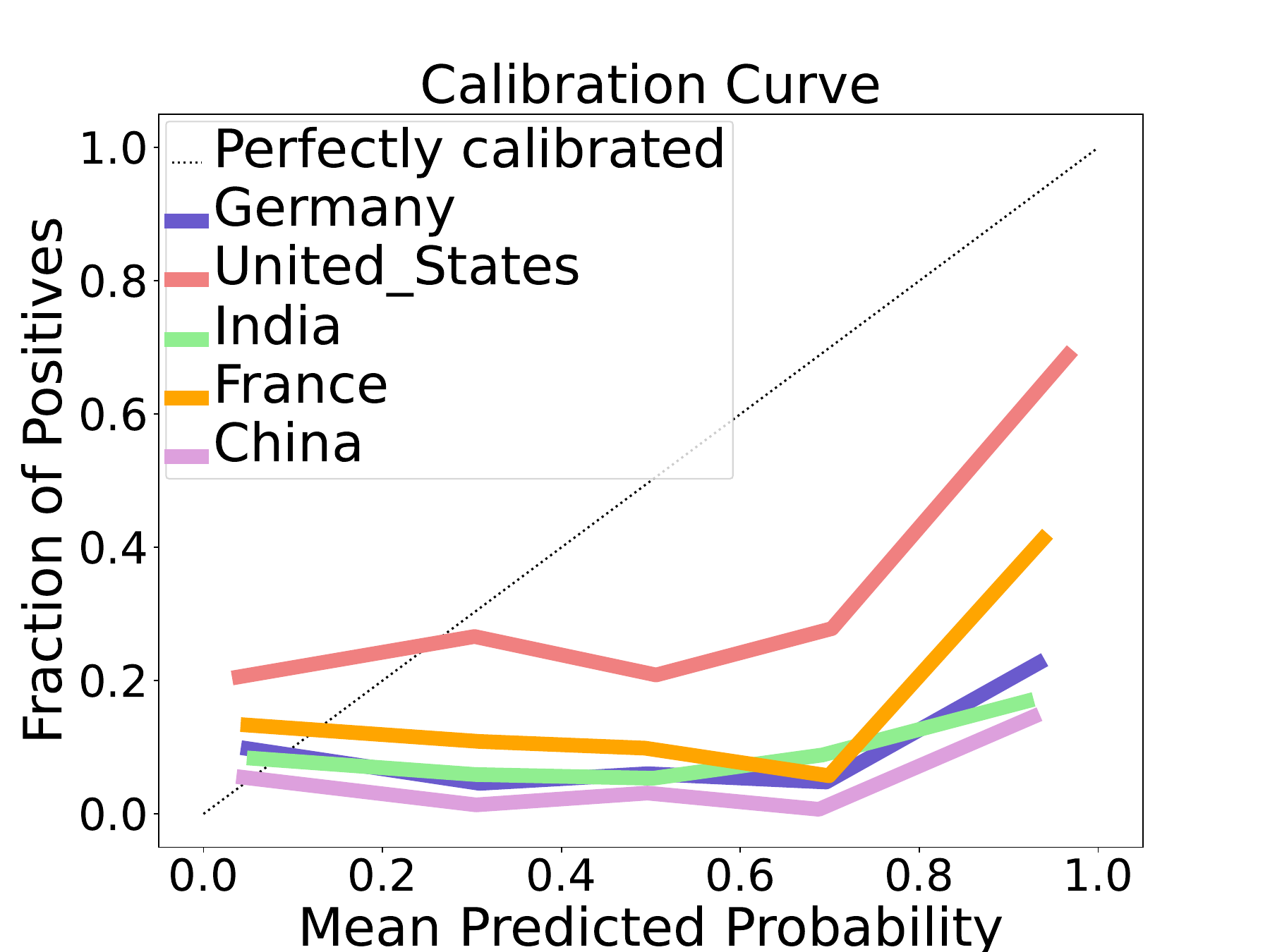} }~\label{fig:nation_mistral_nli_before}}
	\subfloat[\centering After Calibration (\textit{Mistral})]{{\includegraphics[width=0.32\textwidth]{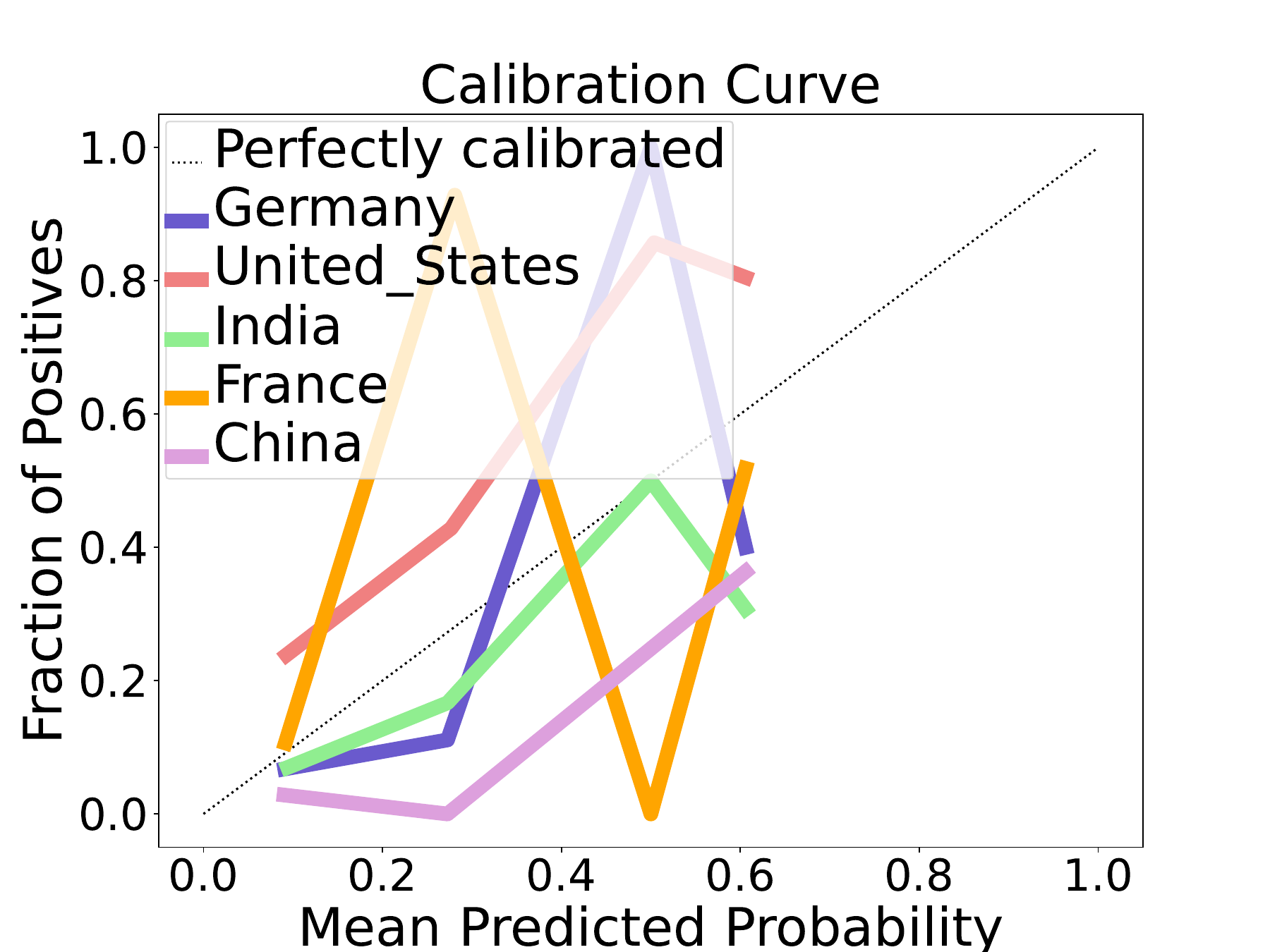} }~\label{fig:nation_mistral_nli_recal}}%
 \subfloat[\centering Ours (\textit{Mistral})]{{\includegraphics[width=0.32\textwidth]{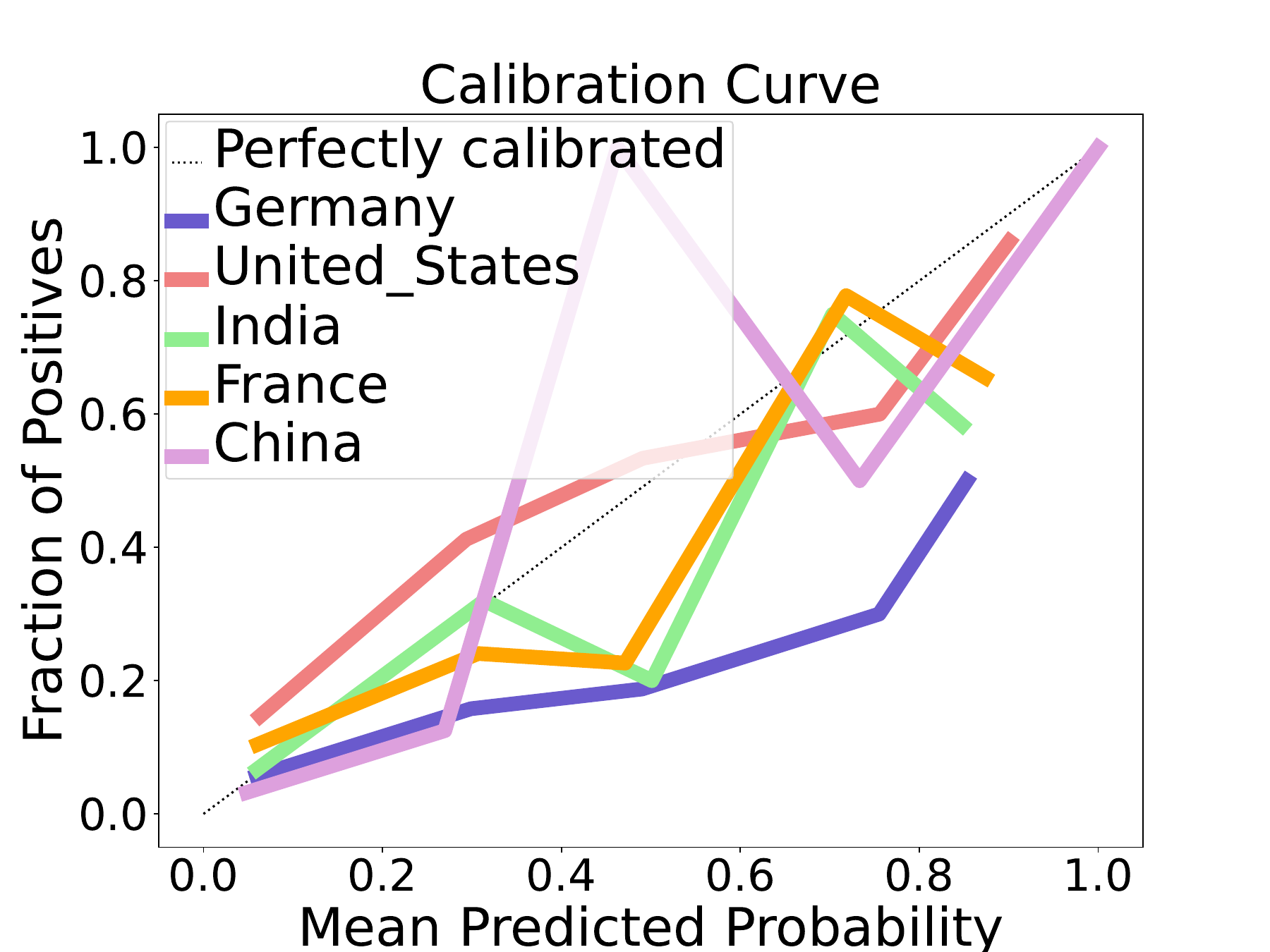} }~\label{fig:nation_mistral_nli_reconf}}%
 
\caption{Comparing calibrations across different nationality groups for the Mistral-7B and LLaMA-7B. We use merged results of three regions. The confidence method here is SelfCheckGPT. }
	\label{fig:recal_ours_nationality}%
 \vspace{-10pt}
\end{figure*}

\paragraph{Calibration Curves.}
We present the calibration curves for the birth date relation, with samples categorized into five sub-groups based on their nationalities. In Figure~\ref{fig:nation_llama7b_nli_before}, it is evident that LLaMA exhibits overconfidence across all nationalities. Following calibration~\ref{fig:nation_llama7b_nli_recal}, there is an improvement for samples with predicted confidence scores less than 0.5, but challenges persist for samples with higher confidences. However, after reconfidencing, as illustrated in Figure~\ref{fig:nation_llama7b_nli_reconf}, the calibration curves demonstrate substantial enhancement, although perfection is not achieved. This observation aligns with similar trends observed in the Mistral model (Figure~\ref{fig:nation_mistral_nli_reconf}).

\begin{figure*}[t]%
	\centering
	\subfloat[\centering  \textit{Recal-LLaMA-JAFC}]{{\includegraphics[width=0.24\textwidth]{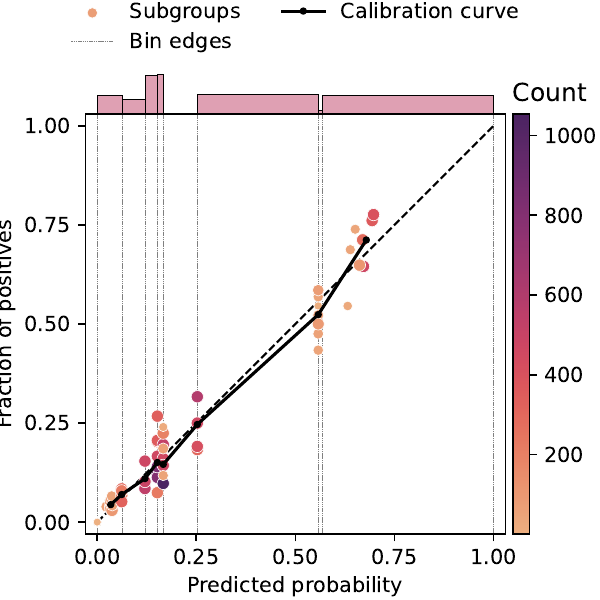} }}
	\subfloat[\centering \textit{Ours-LLaMA-JAFC}]{{\includegraphics[width=0.24\textwidth]{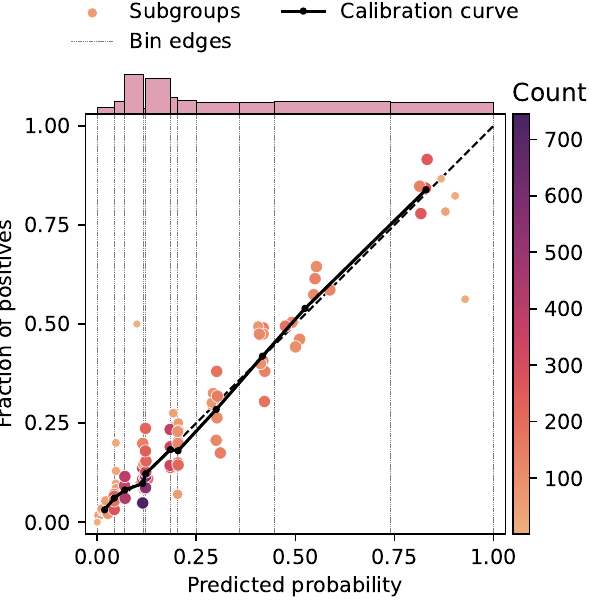} }}%
 \subfloat[\centering \textit{Recal-Mistral-JAFC}]{{\includegraphics[width=0.24\textwidth]{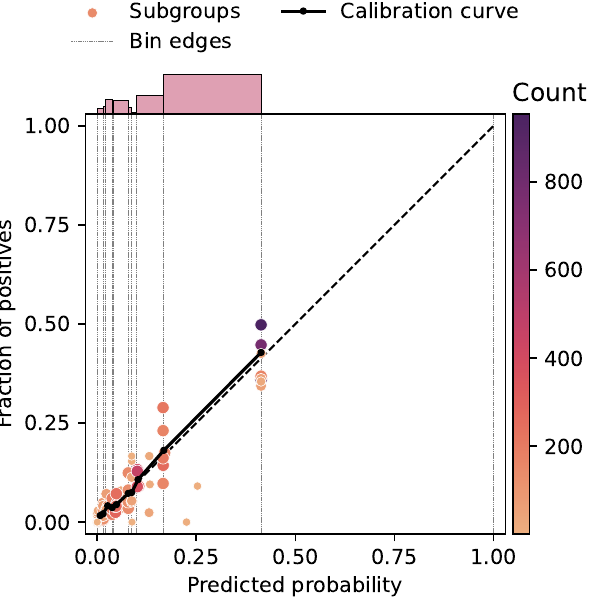} }}%
 \subfloat[\centering \textit{Ours-Mistral-JAFC}]{{\includegraphics[width=0.24\textwidth]{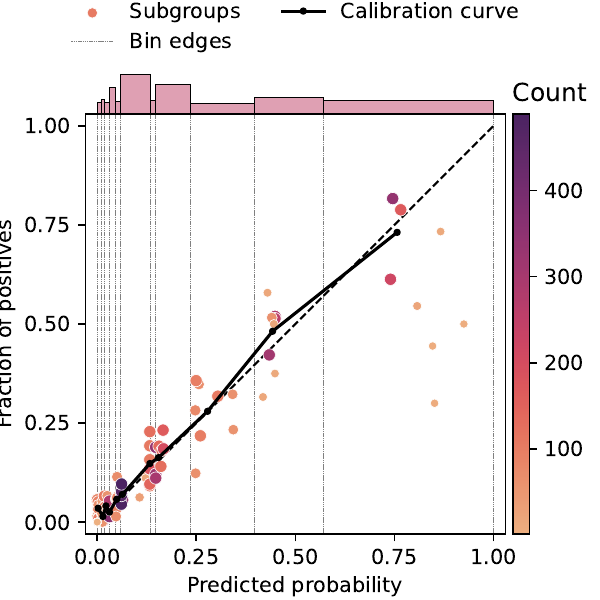} }}%

    \bigskip
\subfloat[\centering  \textit{Recal-LLaMA-SCGPT}]{{\includegraphics[width=0.24\textwidth]{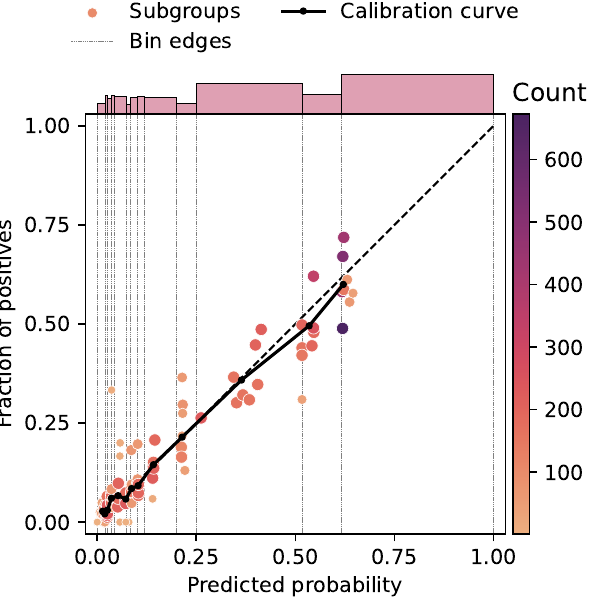} }}
	\subfloat[\centering \textit{Ours-LLaMA-SCGPT}]{{\includegraphics[width=0.24\textwidth]{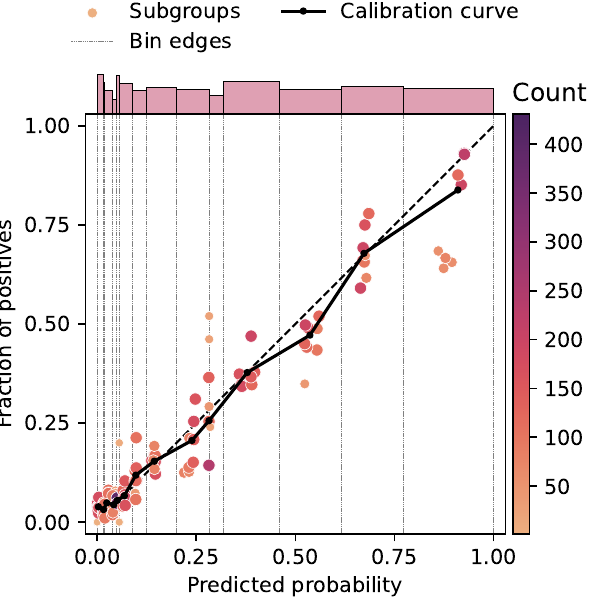} }}%
 \subfloat[\centering \textit{Recal-Mistral-SCGPT}]{{\includegraphics[width=0.24\textwidth]{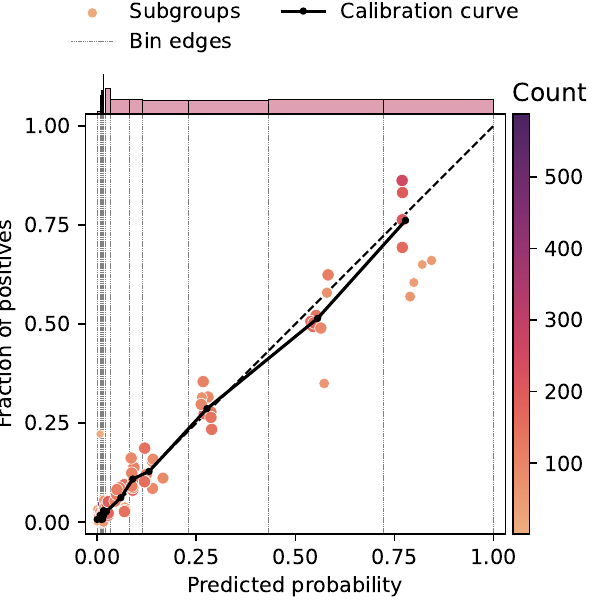} }~\label{fig:physical_mistral_nli_S_after_recal}}%
 \subfloat[\centering \textit{Ours-Mistral-SCGPT}]{{\includegraphics[width=0.24\textwidth]{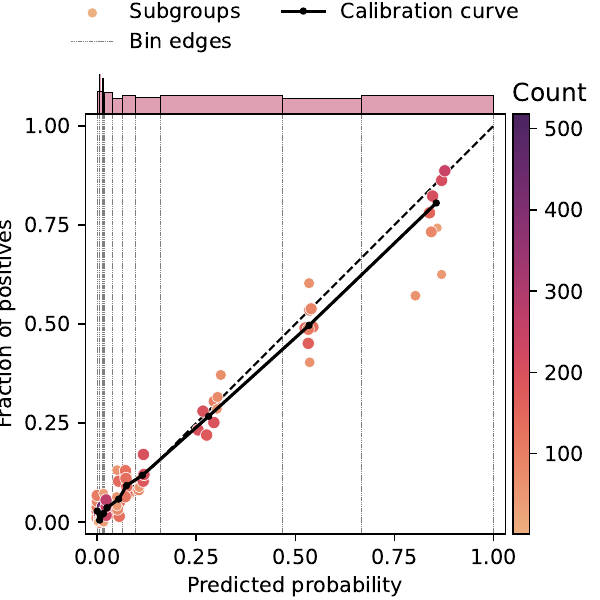} }~\label{fig:physical_mistral_nli_S_after_reconf}}%
 
\caption{Comparing calibrations on popularity groups. Each bin is divided into 8 groups. "Recal" means the Calibration method. }
	\label{fig:latent_recal_ours_popularity}%
 \vspace{-10pt}
\end{figure*}

\paragraph{Grouping Diagrams.}
We illustrate the grouping diagrams for popularity sub-groups, where all samples are evenly distributed into eight partitions based on the number of backlinks. Subsequently, we depict diagrams following calibration and reconfidencing in Figure~\ref{fig:latent_recal_ours_popularity}. In general, when comparing the calibration method to reconfidencing, the latter exhibits superior calibration of confidence scores. For instance, in Figure~\ref{fig:physical_mistral_nli_S_after_reconf}, the calibration curve appears more diagonal compared to Figure~\ref{fig:physical_mistral_nli_S_after_recal}, indicating improved calibration through reconfidencing.

Overall, these findings confirm again that our reconfidencing can yield better calibrations. 

\subsection{Experiments on Open-ended QA Tasks}\label{sec:open_qa}
Since our method reduce the grouping loss for entity-based queries, one may ask can our reconfidencing method be applied for other datasets or open-ended generation tasks. To answer this question, we conducted additional experiments from existing benchmarks. We follow the setting in this ~\citet{manakul2023selfcheckgpt} to conduct experiments on three QA datasets: SciQ~\cite{welbl2017crowdsourcing}, TriviaQ~\cite{joshi2017triviaqa} and Truthful QA~\cite{lin2022truthfulqa}. 
Besides, we include another open-ended generation task from the medical domain, Medical QA\footnote{\url{https://huggingface.co/datasets/medalpaca/medical\_meadow\_medical\_flashcards}}. 
Some details of the four QA datasets are shown in the Table~\ref{tab:qa_dataset_stats}. As for evaluation, we use the API of GPT-3.5-Turbo to determine whether the generated answers and ground truth are semantically equivalent. The LLM to generate confidence scores here is LLaMA-13B.

The experimental results are shown in Table~\ref{tab:reconfidencing_results_qa}. 
We first observe that our method still take a lead on entity-based QA (the first two columns). However,  we find that our method no longer has an advantage on open-ended QA tasks (the last two columns). 

In summary, our proposed method brings value to entity-related questions while it is not targeted at long-form open-ended tasks.

\subsection{Experiments on Other Relations}
\begin{figure*}[tb]%
\centering
\subfloat[\centering Before (\textit{Mistral})]{{\includegraphics[width=0.32\textwidth]{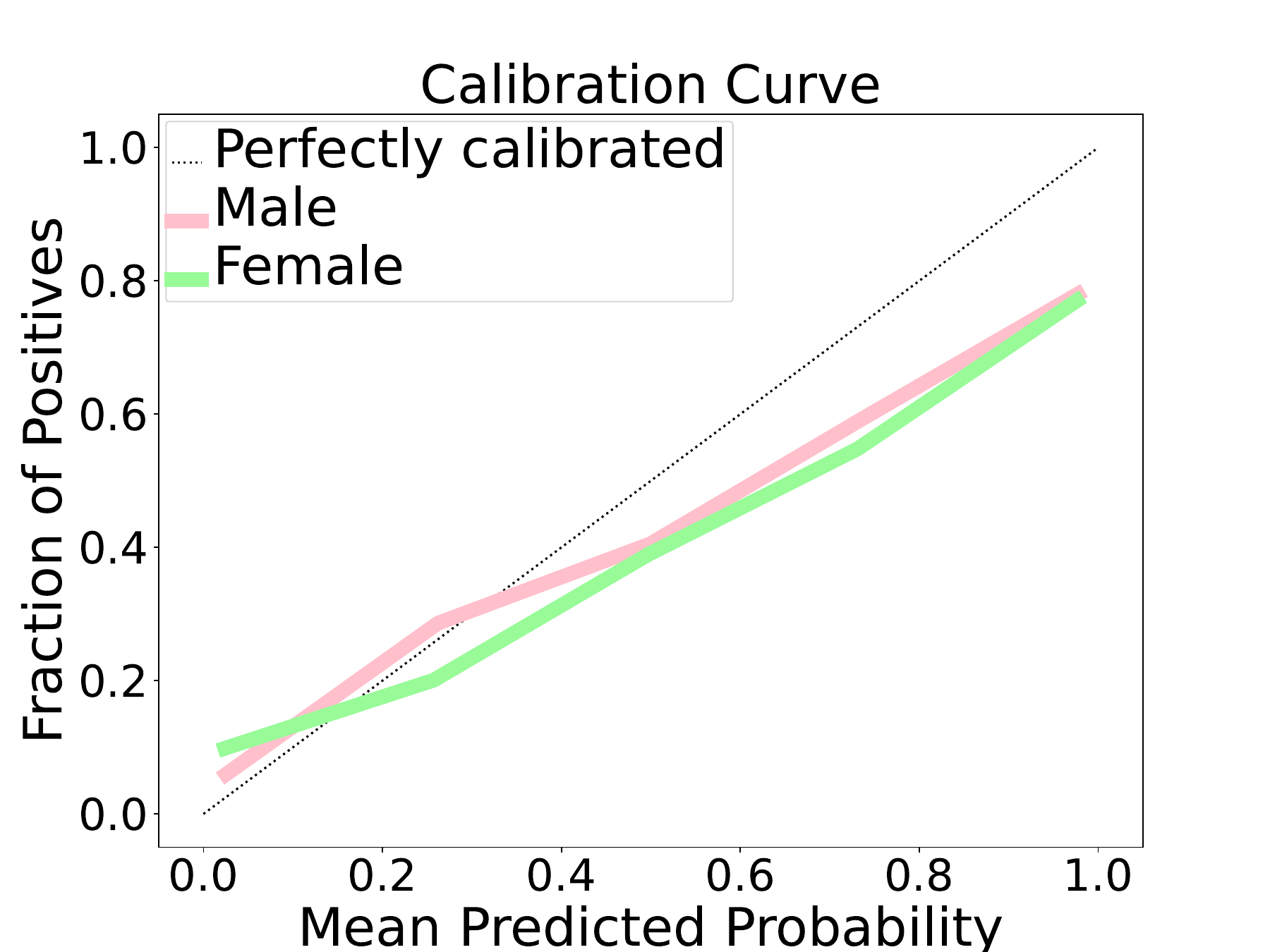}}}
	\subfloat[\centering After Calibration (\textit{Mistral})]{{\includegraphics[width=0.32\textwidth]{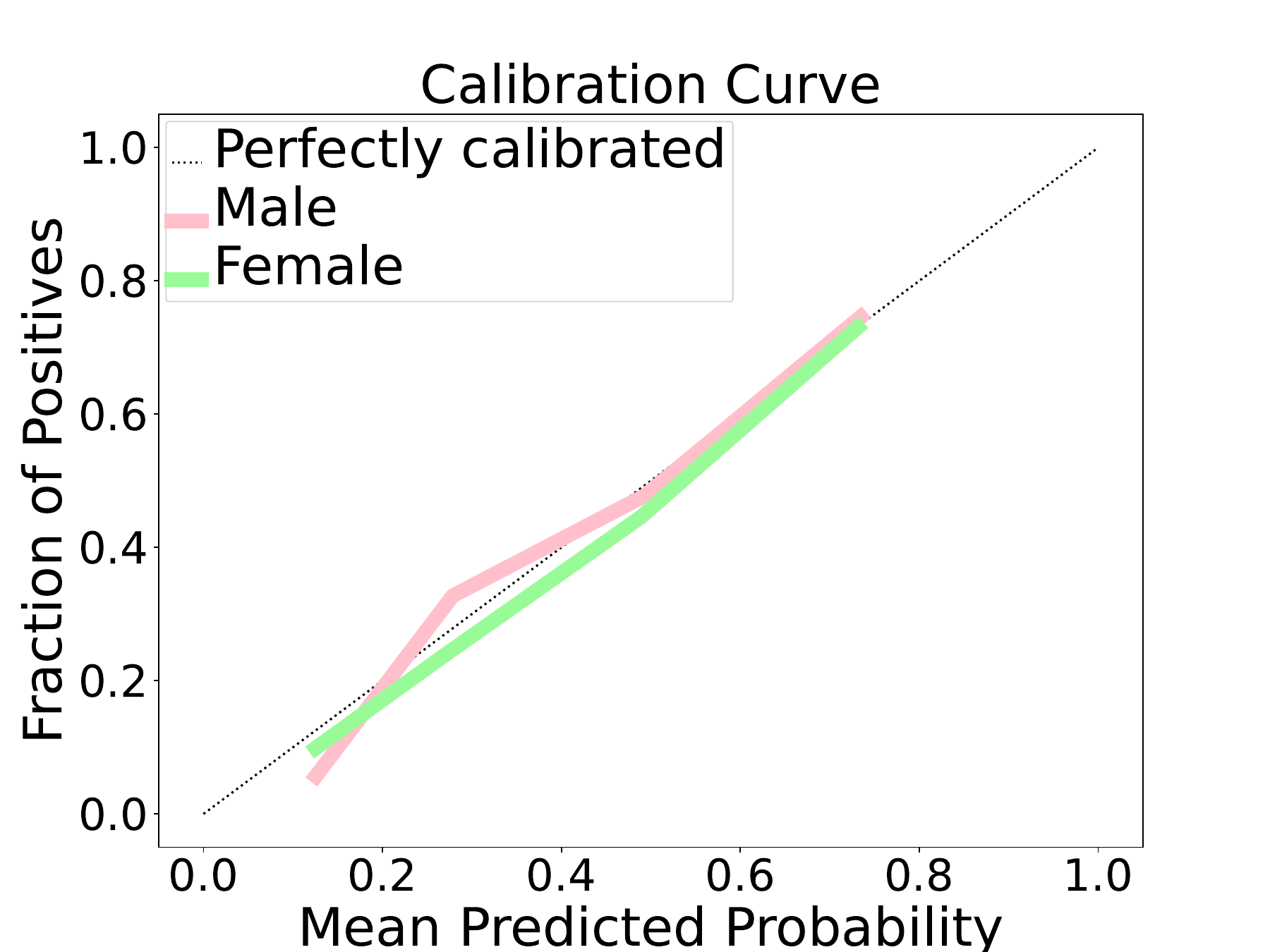} }}%
 \subfloat[\centering Ours (\textit{Mistral})]{{\includegraphics[width=0.32\textwidth]{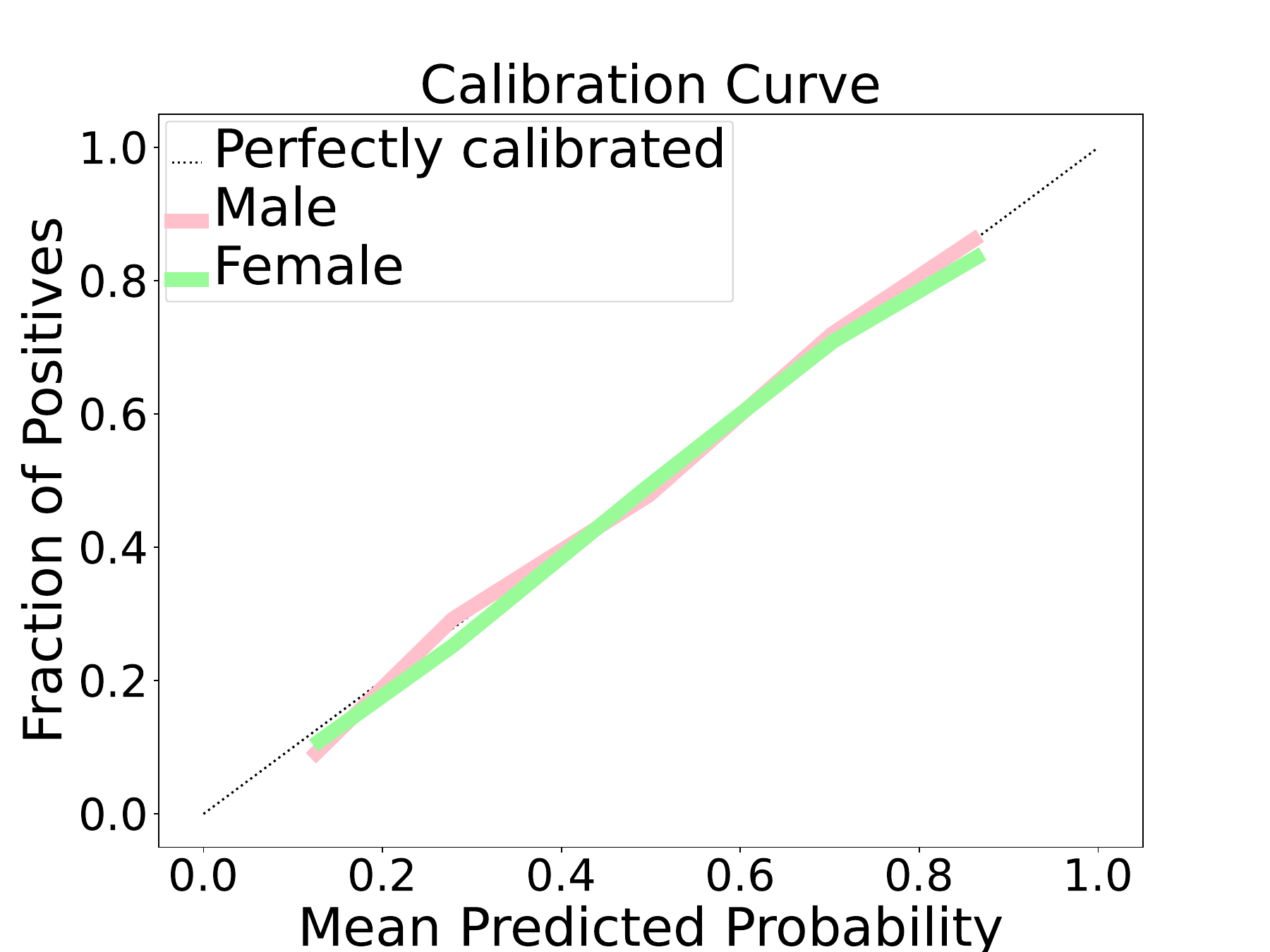} }}%

\caption{Comparing calibrations across different gender groups of the \texttt{Birth Place} relation for the Mistral-7B. The confidence method here is SelfCheckGPT. }
	\label{fig:gender_group}%
 \vspace{-10pt}
\end{figure*}

\begin{figure*}[t]%
\centering
\subfloat[\centering Before (\textit{Mistral})]{{\includegraphics[width=0.32\textwidth]{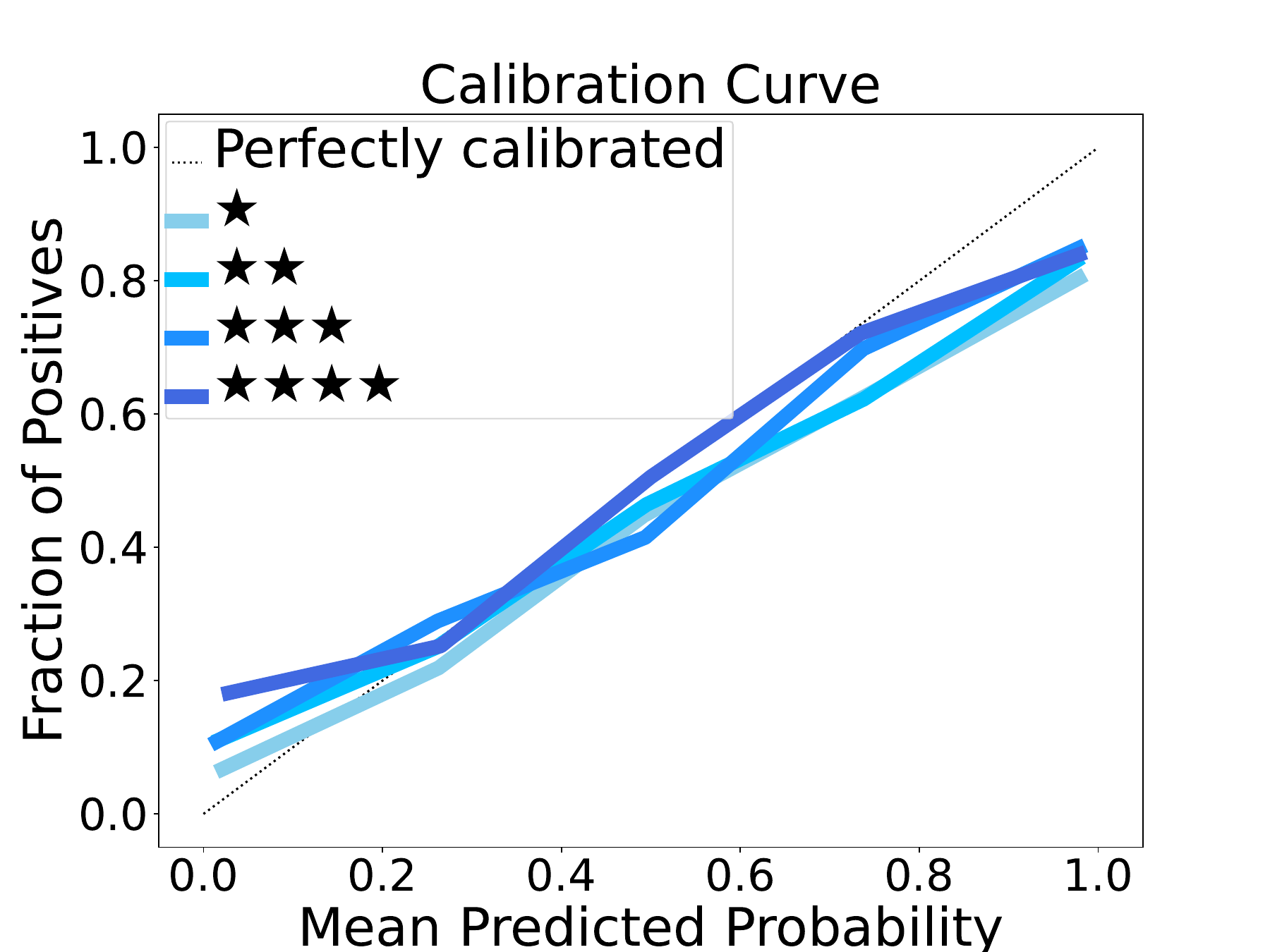}}}
	\subfloat[\centering After Calibration (\textit{Mistral})]{{\includegraphics[width=0.32\textwidth]{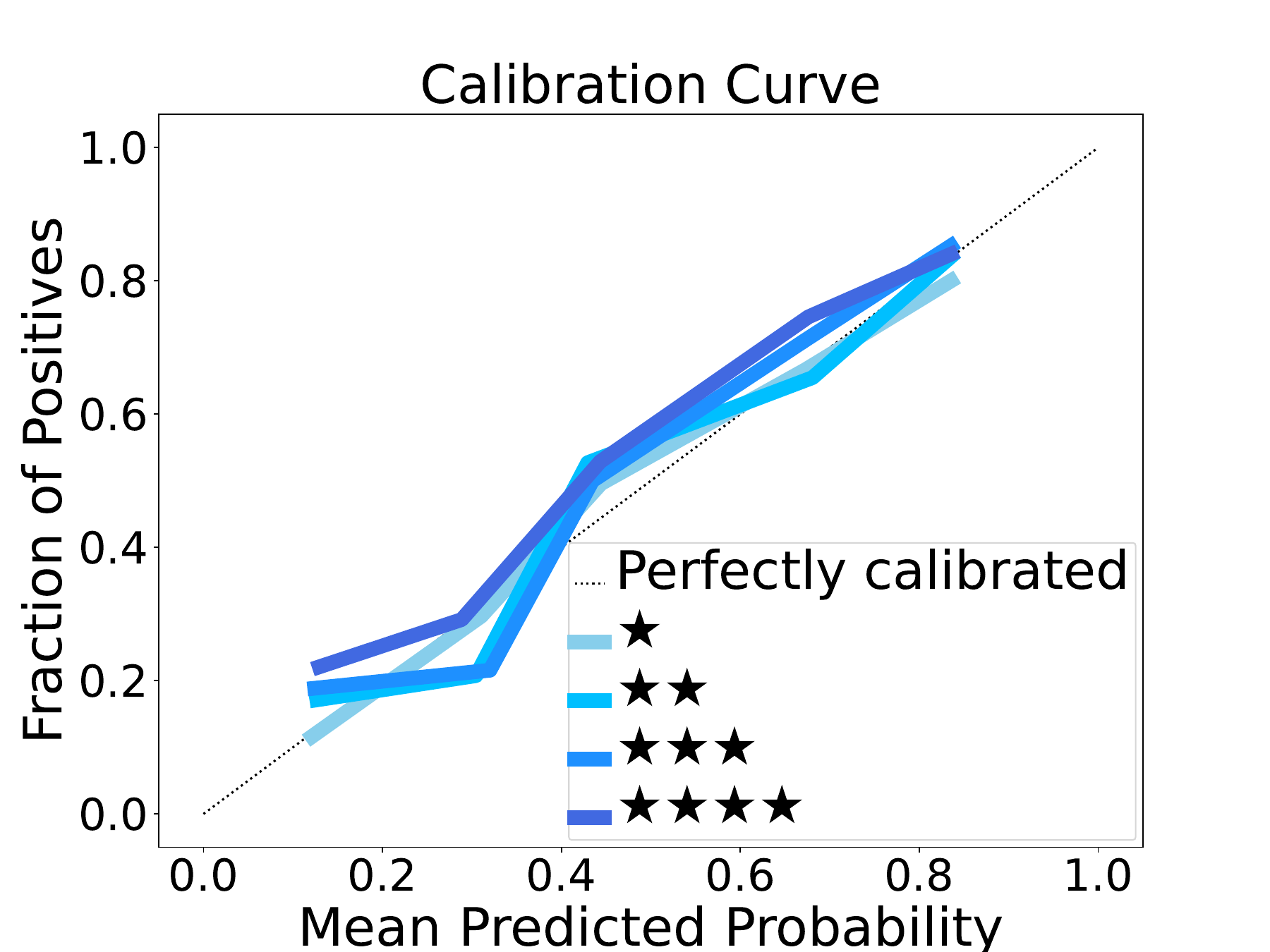} }}%
 \subfloat[\centering Ours (\textit{Mistral})]{{\includegraphics[width=0.32\textwidth]{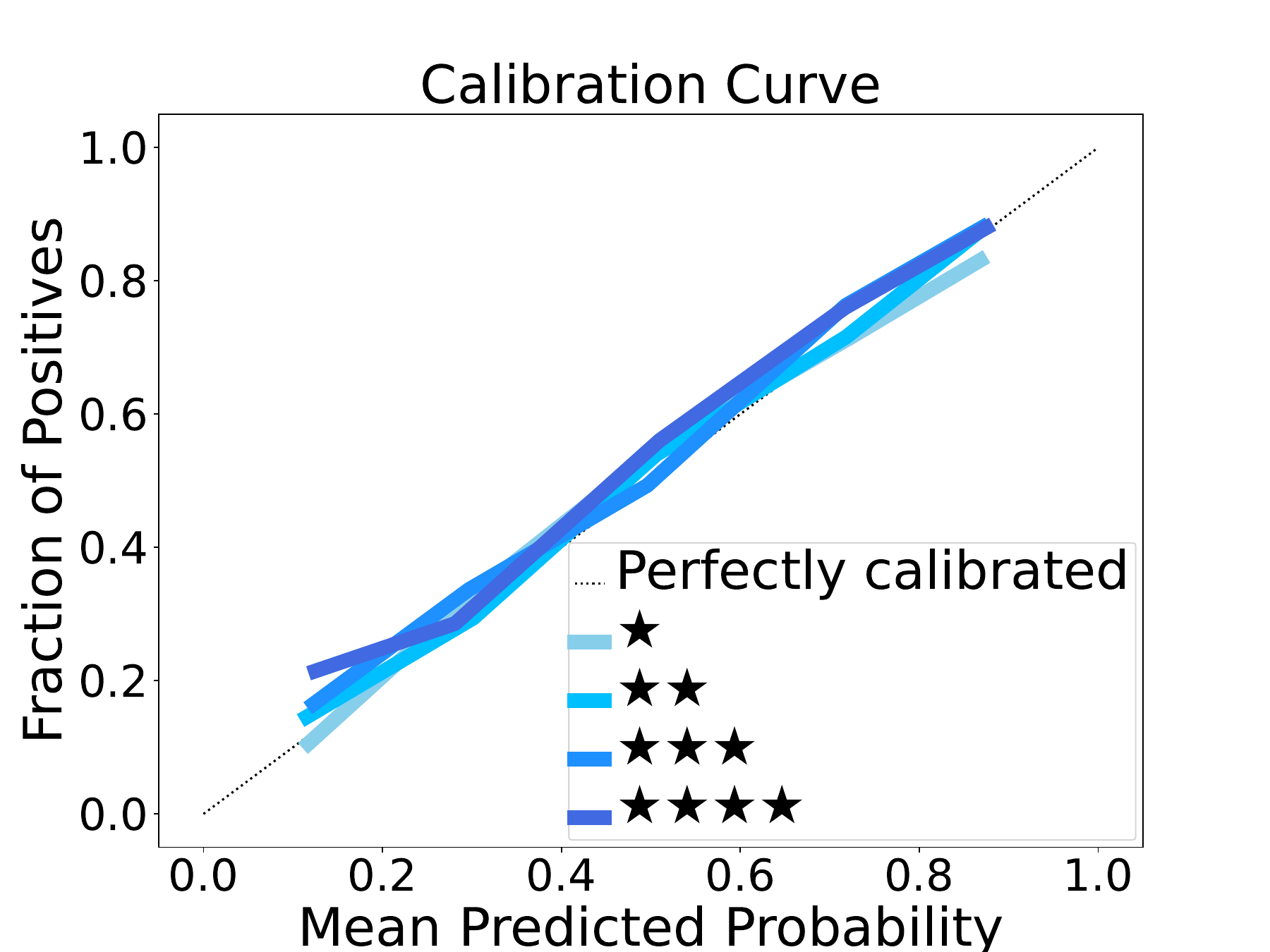} }}%
 
\caption{Comparing calibrations across different popularity groups of the \texttt{LocationCreated} relation for the Mistral-7B. The confidence method here is SelfCheckGPT. }
	\label{fig:location_created}%
 \vspace{-10pt}
\end{figure*}

To show the scalability of our reconfidencing method, we conduct experiments on another two relations: \texttt{Birth\_Place} and \texttt{LocationCreated}. To study the fairness of LLMs better, we introduce gender groups in the \texttt{Birth\_Place} dataset. 
In Figure~\ref{fig:gender_group}, we draw curves of \texttt{Birth\_Place} for both male and female sub-groups. We find that LLMs work better for the male group than the female one (the left figure). Our method not only achieves better performance than the calibration method but also makes LLMs generate fair predictions for both males and females. 
In figure~\ref{fig:location_created}, we also draw the calibration curves for the \texttt{LocationCreated} relation (a film is created in which country). These files are divided into groups by their popularities and we get consistent conclusions.

\subsection{The Impact of Partition Numbers}\label{sec:partition_number}
To study the impact of the granularity of partition, we vary the number of partitions for LLaMA-13B and check the performances. The results are shown in Table~\ref{tab:partition_number}. If there are too few partitions ($p \leq 4$), it will decrease the performance of our method. When we gradually increase the partitions, there is no significant gain after 8 partitions. In our paper, the partition number is 8 for all datasets.

\begin{table}[bt] 
	\centering
        \small
	\setlength{\tabcolsep}{4.0mm}{
		\begin{threeparttable} 
			\begin{tabular}{c|ccc}  
				\toprule
				\textbf{Method}
				&\multicolumn{3}{c}{\textbf{\underline{\texttt{Composer}}}}\cr
			&Brier~\textcolor{brandeisblue}{$\downarrow$}&CL~\textcolor{brandeisblue}{$\downarrow$}&GL~\textcolor{brandeisblue}{$\downarrow$}\cr
			 \midrule
            Before&68.16&37.89&3.05\cr
            Calibration &30.62&0.31&3.6\cr
            Ours (p=2)&26.52&0.76&1.04\cr
            Ours (p=4)&26.12&0.62&0.0\cr
            Ours (p=8)&26.01&0.54&0.0\cr
            Ours (p=16)&25.87&0.56&0.37\cr
            Ours (p=32)&25.44&0.72&0.0\cr
            Ours (p=64)&25.9&1.32&0.0\cr
				\bottomrule
			\end{tabular}
			\caption{Evaluating calibration of various confidence methods. Here, we compare \textit{Just Asking for Calibration (JAFC)}~\cite{tian2023just} and \textit{SelfCheckGPT~\cite{manakul2023selfcheckgpt}}. CL and GL mean calibration loss and grouping loss, respectively. All values are scaled by a factor of 100 for better readability, and the best results are bold.}%
			\label{tab:partition_number}%
			\vspace{-5pt}
		\end{threeparttable} 
	}
\end{table}

\begin{table*}[!t]  
	\centering  
	\footnotesize
	\begin{threeparttable} 
		\begin{tabular}{c|c|p{2cm}|p{4cm}|p{4cm}}  
			\toprule  
			Relation&Size&Head&Query Example&Answer Example\cr
			\midrule
			SciQ&1,000&Scientific entity-based QA&\textit{What tells you how much of the food you should eat to get the nutrients listed on the label?} &\textit{serving size}\cr 
            \midrule
            TriviaQ&1,000&General entity-based QA&\textit{Which American-born Sinclair won the Nobel Prize for Literature in 1930?} &\textit{Sinclair Lewis}\cr 
            \midrule
            Truthful QA&817&Commonsense open-ended QA&\textit{What happens to you if you eat watermelon seeds?} &\textit{The watermelon seeds pass through your digestive system}\cr 
            \midrule
            Medical QA&1,000&Medical open-ended QA&\textit{What is the relationship between very low Mg2+ levels, PTH levels, and Ca2+ levels?} &\textit{Very low Mg2+ levels correspond to low PTH levels which in turn results in low Ca2+ levels.}\cr 
			\bottomrule  
		\end{tabular}
		\caption{Description of four QA evaluation dataset. We follow the setting in this paper (https://aclanthology.org/2023.emnlp-main.330/)
        to run experiments. Medical QA is adapted from the 
        \texttt{medical\_meadow\_medical\_flashcards} on HuggingFace Datasets.   As for evaluation, we use the API of GPT-3.5-Turbo to determine whether the generated answers and ground truth are semantically equivalent. The LLM here is LLaMA-13B.
 }\label{tab:qa_dataset_stats}
	\end{threeparttable}  
\end{table*}

\begin{table*}[tb] 
	\centering
	\setlength{\tabcolsep}{1.2mm}{
		\begin{threeparttable} 
			\begin{tabular}{c|ccc|ccc|ccc|ccc}  
				\toprule
				\textbf{Method}
				&\multicolumn{3}{c|}{\textbf{\underline{\texttt{SciQ}}}}&\multicolumn{3}{c|}{\textbf{\underline{\texttt{TriviaQ}}}}&\multicolumn{3}{c|}{\textbf{\underline{\texttt{Truthful\_QA}}}}&\multicolumn{3}{c}{\textbf{\underline{\texttt{Medical\_QA}}}}\cr
			&Brier~\textcolor{brandeisblue}{$\downarrow$}&CL~\textcolor{brandeisblue}{$\downarrow$}&GL~\textcolor{brandeisblue}{$\downarrow$}&Brier~\textcolor{brandeisblue}{$\downarrow$}&CL~\textcolor{brandeisblue}{$\downarrow$}&GL~\textcolor{brandeisblue}{$\downarrow$}&Brier~\textcolor{brandeisblue}{$\downarrow$}&CL~\textcolor{brandeisblue}{$\downarrow$}&GL~\textcolor{brandeisblue}{$\downarrow$}&Brier~\textcolor{brandeisblue}{$\downarrow$}&CL~\textcolor{brandeisblue}{$\downarrow$}&GL~\textcolor{brandeisblue}{$\downarrow$}\cr
\multicolumn{13}{c}{\cellcolor{gray!15} \textit{LLaMA-13B-SelfCheckGPT}}\cr			          
Before &94.83&52.53&5.19&64.96&17.91&0.0&95.14&61.07&1.17&99.44&70.21&0.0\cr
Calibration &50.16&3.3&2.74&51.43&4.47&0.0&38.9&3.27&0.0&29.62&1.39&0.0\cr
Ours &\colorbox{blue!20}{48.65}&\colorbox{red!20}{5.15}&\colorbox{blue!20}{0.0}&\colorbox{blue!20}{51.0}&\colorbox{red!20}{6.92}&\colorbox{blue!20}{0.0}&\colorbox{red!20}{41.36}&\colorbox{red!20}{7.06}&\colorbox{red!20}{0.0}&\colorbox{red!20}{32.58}&\colorbox{red!20}{3.52}&\colorbox{red!20}{0.32}\cr
    \bottomrule
			\end{tabular}
			\caption{Comparing methods on four QA tasks of after calibration and our reconfidencing.  Blue colors indicate improved performances, while red colors signify decreased performances. All values are scaled by a factor of 100 for better readability.}%
			\label{tab:reconfidencing_results_qa}%
			\vspace{-5pt}
		\end{threeparttable} 
	}
\end{table*}

\begin{figure*}[t]%
\centering

 \subfloat[\centering Before (\textit{LLaMA-7B})]{{\includegraphics[width=0.32\textwidth]{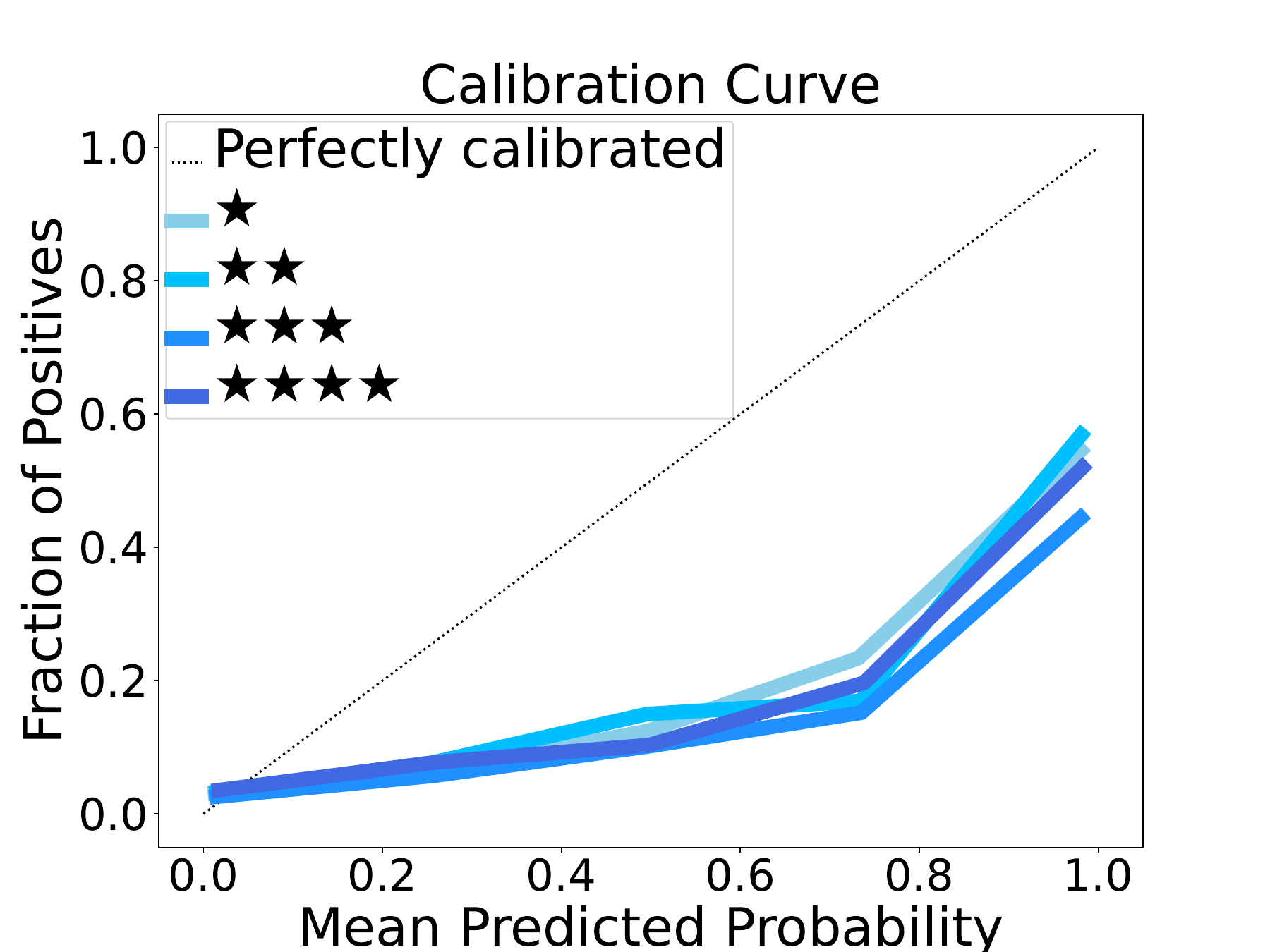} }\label{fig:llama7b_nli_before}}
	\subfloat[\centering After Calibration (\textit{LLaMA-7B})]{{\includegraphics[width=0.32\textwidth]{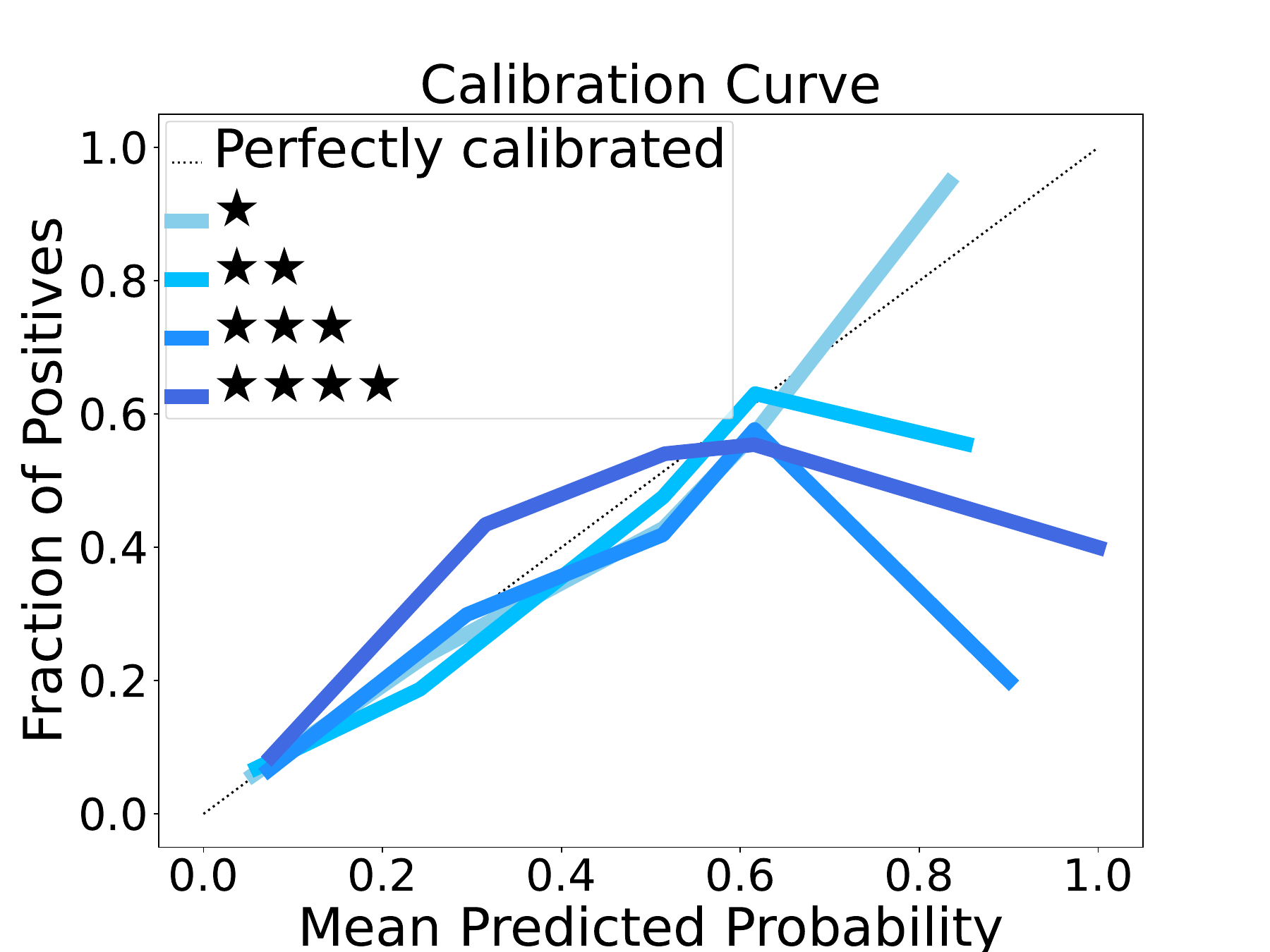} }\label{fig:llama7b_nli_recal}}%
 \subfloat[\centering Ours (\textit{LLaMA-7B})]{{\includegraphics[width=0.32\textwidth]{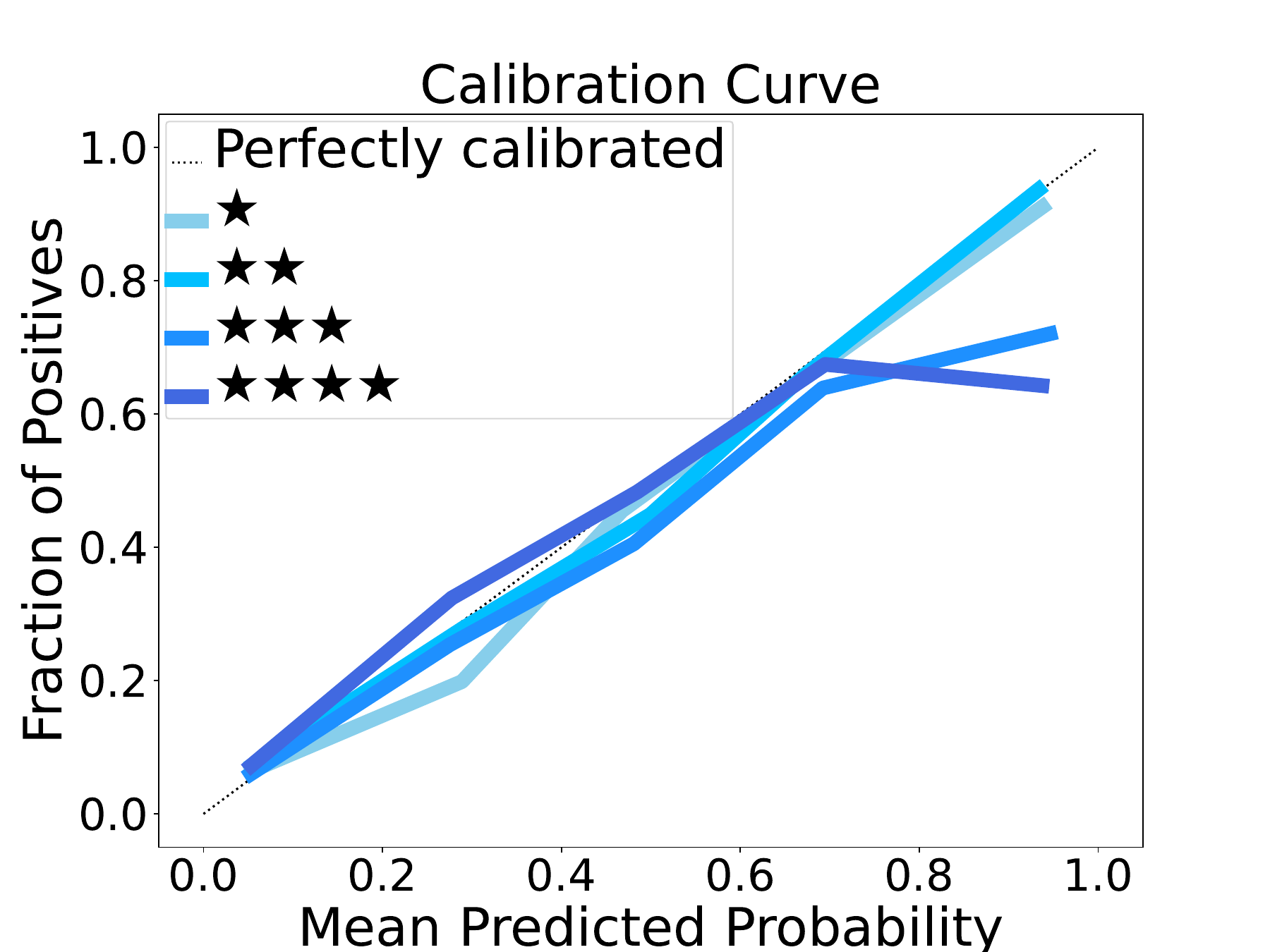} }\label{fig:llama7b_nli_reconf}}

 \bigskip

\subfloat[\centering Before (\textit{LLaMA-13B})]{{\includegraphics[width=0.32\textwidth]{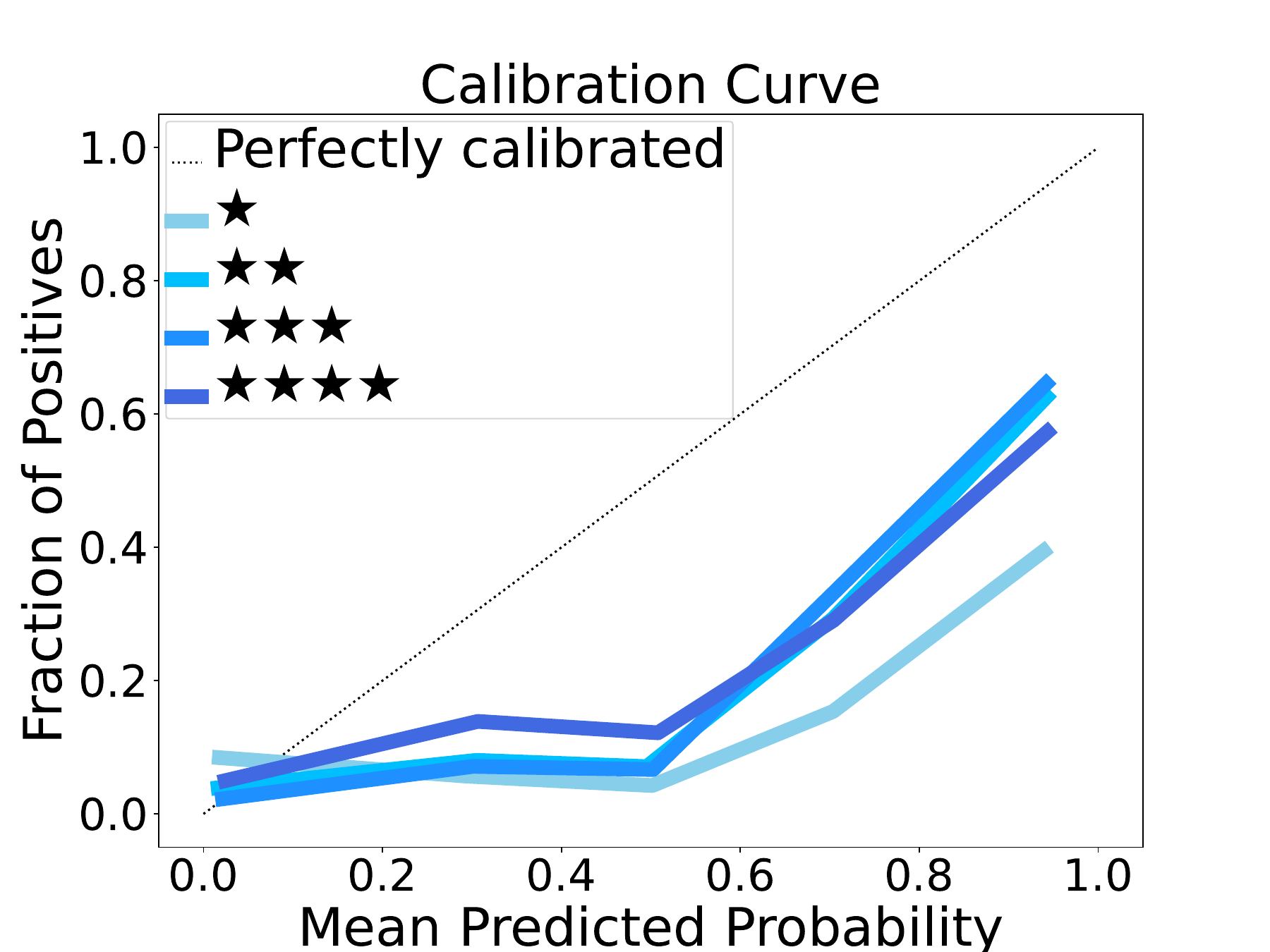}}}
	\subfloat[\centering After Calibration (\textit{LLaMA-13B})]{{\includegraphics[width=0.32\textwidth]{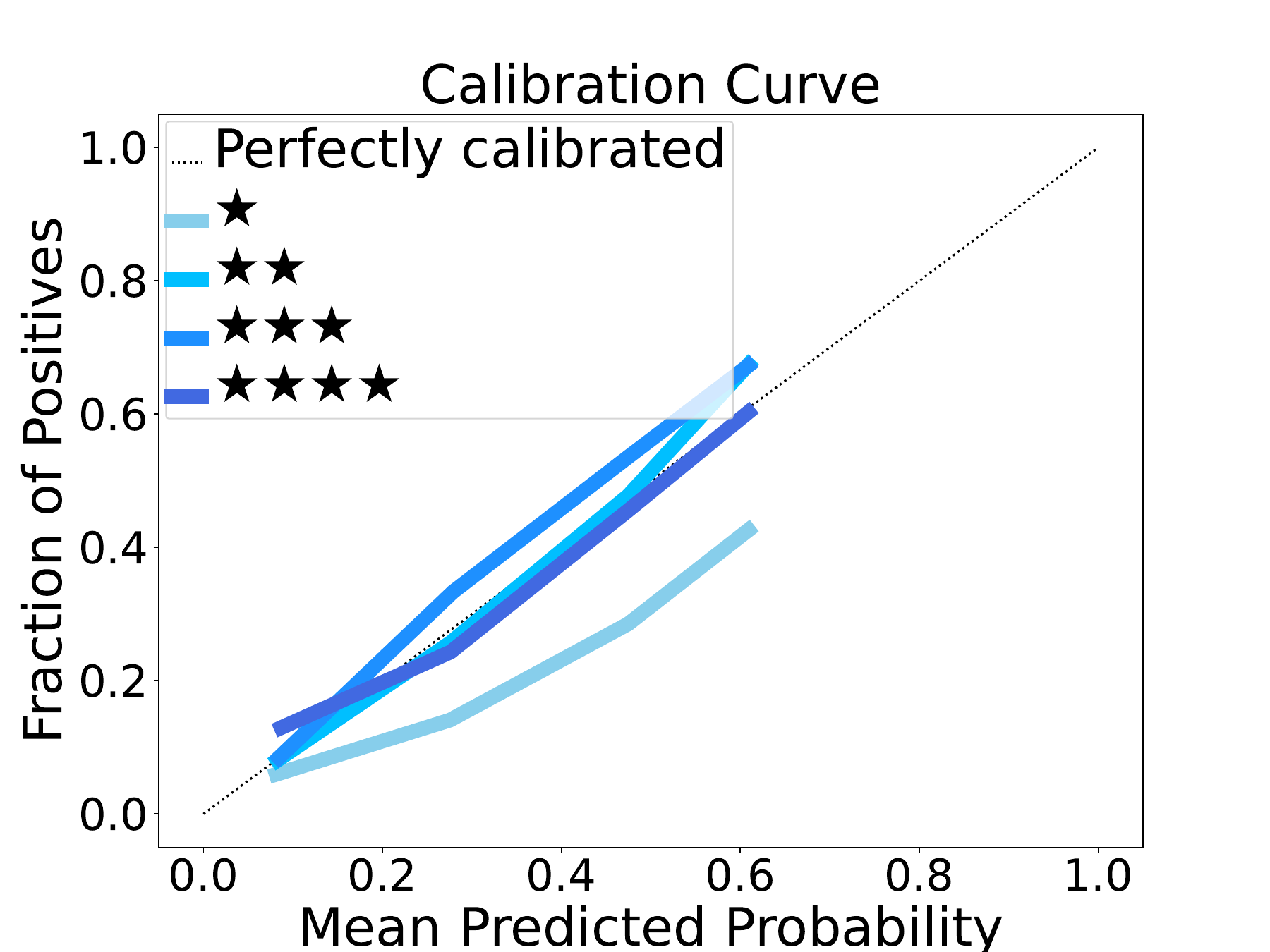} }}%
 \subfloat[\centering Ours (\textit{LLaMA-13B})]{{\includegraphics[width=0.32\textwidth]{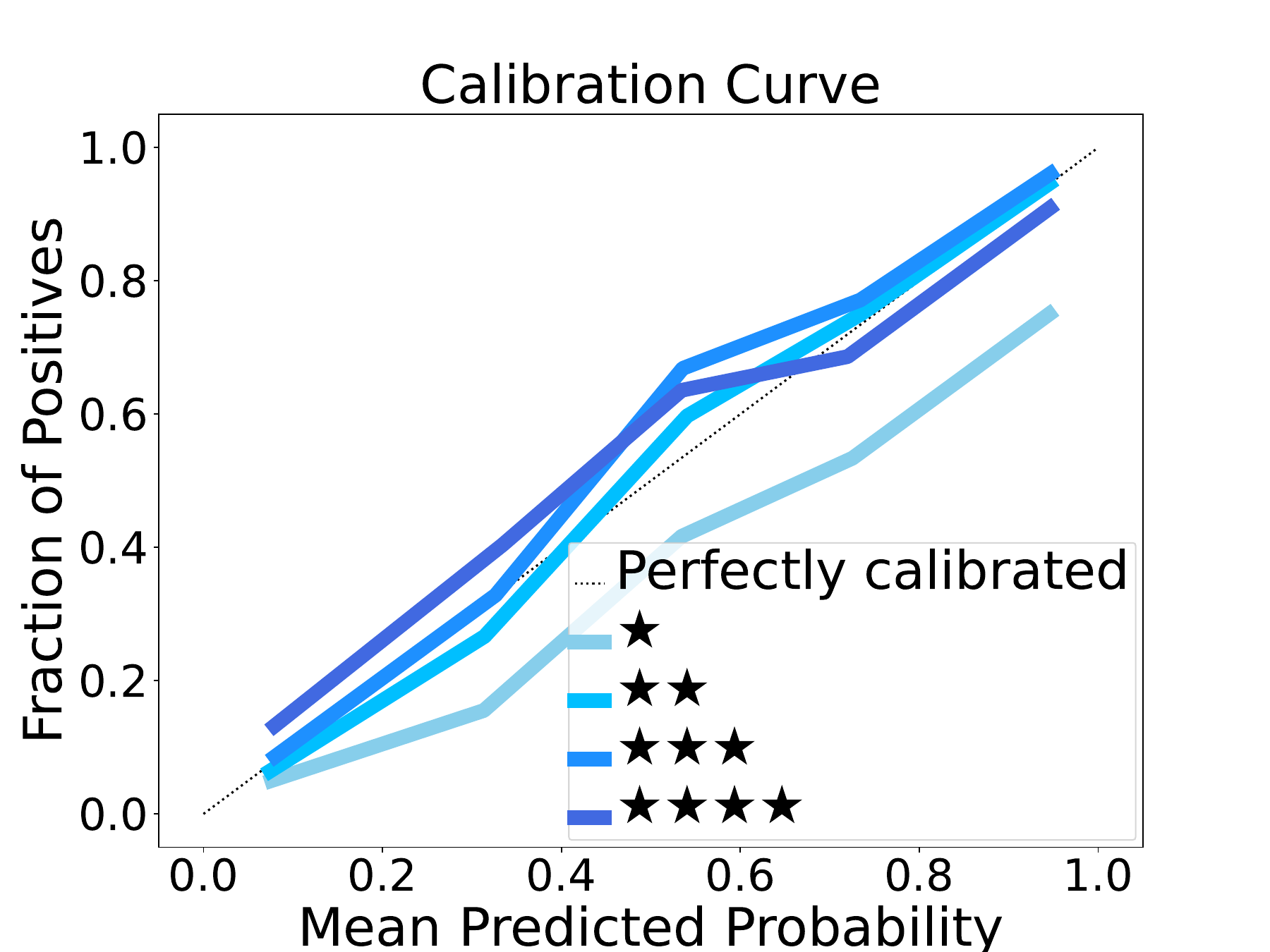} }}%
 
\caption{Comparing calibrations across different popularity groups of the \texttt{Birth Date} relation for the LLaMA-13B. The confidence method here is SelfCheckGPT. }
	\label{fig:popularity_7b_13b}%
 \vspace{-10pt}
\end{figure*}

\end{document}